%% file: main.tex
\documentclass[10pt,journal,compsoc]{IEEEtran}

\usepackage{tikz}
\usepackage{pgfplots}
\pgfplotsset{compat=1.18}
\usetikzlibrary{arrows.meta, shapes, calc}

\usepackage{cite}
\usepackage{multirow} 
\usepackage{amsmath,amsfonts}
\usepackage{subfig}     
\usepackage{caption}
\usepackage{array}
\usepackage{amsmath,amssymb,amsfonts}
\usepackage{algorithmic}
\usepackage{graphicx}
\usepackage{textcomp}
\usepackage{xcolor}
\usepackage{subcaption}
\usepackage{url}  
\usepackage[linesnumbered,ruled,vlined]{algorithm2e} 
\usepackage{subfig}
\usepackage{textcomp}
\usepackage{stfloats}
\usepackage{url}
\usepackage{verbatim}
\usepackage{graphicx}
\usepackage{cite}
\begin{document}

\title{IVGAE: Handling Incomplete Heterogeneous Data with a Variational Graph Autoencoder}

\author{
    Youran Zhou, 
    Mohamed Reda Bouadjenek, 
    and Sunil Aryal%
    \thanks{
        School of Information Technology, Deakin University, Victoria, Australia.
        E-mail: \{echo.zhou, reda.bouadjenek, sunil.aryal\}@deakin.edu.au
    }
}
\markboth{Journal of \LaTeX\ Class Files,~Vol.~14, No.~8, August~2021}%
{Shell \MakeLowercase{\textit{et al.}}: A Sample Article Using IEEEtran.cls for IEEE Journals}

\IEEEpubid{0000--0000/00\$00.00~\copyright~2021 IEEE}

\maketitle


\maketitle

\begin{abstract}
Handling missing data remains a fundamental challenge in real-world tabular datasets, especially when data are heterogeneous with both numerical and categorical features. Existing imputation methods often fail to capture complex structural dependencies and handle heterogeneous data effectively. We present \textbf{IVGAE}, a Variational Graph Autoencoder framework for robust imputation of incomplete heterogeneous data. IVGAE constructs a bipartite graph to represent sample-feature relationships and applies graph representation learning to model structural dependencies. A key innovation is its \textit{dual-decoder architecture}, where one decoder reconstructs feature embeddings and the other models missingness patterns, providing structural priors aware of missing mechanisms. To better encode categorical variables, we introduce a Transformer-based heterogeneous embedding module that avoids high-dimensional one-hot encoding. Extensive experiments on 16 real-world datasets show that IVGAE achieves consistent improvements in RMSE and downstream F1 across  MCAR, MAR, and MNAR missing scenarios under 30\% missing rates. Code and data are available at: \url{https://github.com/echoid/IVGAE}.
\end{abstract}


\section{Introduction}
\IEEEPARstart{I}{ncomplete} data is an enduring obstacle in real-world machine learning applications, emerging from diverse causes such as sensor malfunctions, human errors, and privacy-driven data omissions. It not only introduces statistical bias but also compromises the reliability of learned models in critical domains such as healthcare, finance, and recommendation systems. This challenge is particularly pronounced in \textit{heterogeneous tabular data}, where both numerical and categorical features coexist and interact in complex ways. Compounding this difficulty is the fact that the \textit{missingness mechanism}—whether data are missing completely at random (MCAR), at random (MAR), or not at random (MNAR)—is often unknown and difficult to infer. These mechanisms fundamentally determine the feasibility and bias of imputation, yet most existing models simplify them under MCAR assumptions. The combination of heterogeneity and mechanism complexity thus makes incomplete data imputation particularly challenging.

Traditional imputation approaches, including statistical heuristics and classical machine learning methods, often fail to capture high-order dependencies or rely on rigid distributional assumptions. For instance, mean or mode~\cite{mean,mean2} imputation operates independently per feature, while $k$-nearest neighbors~\cite{knn}, MICE~\cite{mice}, and matrix factorization~\cite{MF} impose linearity or independence assumptions that generalize poorly to unseen data. Deep generative models, such as GAN-based~\cite{yoon2018gain,misGAN} and VAE-based~\cite{vaem,hivae,missingvae,miwae,notmiwae} frameworks, have shown remarkable progress by learning conditional data distributions. However, they are typically optimized for continuous data and require extensive preprocessing of categorical variables, leading to information loss and suboptimal performance on heterogeneous datasets.

Graph-based approaches have recently emerged as a promising paradigm for imputing structured data, as they naturally capture relational dependencies among samples and features. Representing tabular data as a bipartite graph allows models to jointly learn local interactions and global structural patterns~\cite{grape,igrm}. Graph Neural Networks (GNNs) have been leveraged to enhance imputation quality~\cite{tabgnn,ginn,graphVAE,graphvae2,GraphSAGE}, but existing graph-based imputers are often limited to homogeneous data and tend to overlook the underlying missingness mechanisms, thereby reducing their robustness in MAR and MNAR settings.

These limitations motivate the development of a unified model that can \textbf{(1)} robustly impute missing values in heterogeneous tabular data and \textbf{(2)} explicitly account for complex, mechanism-driven missingness. To this end, we propose \textbf{IVGAE}, an \textbf{I}ncomplete \textbf{V}ariational \textbf{G}raph \textbf{A}uto\textbf{E}ncoder for robust imputation in heterogeneous tabular datasets with structured missingness. IVGAE models the data as a bipartite graph and introduces a \textit{dual-decoder architecture}, where one decoder reconstructs feature embeddings while the other estimates missingness masks as edge probabilities. This mechanism-aware design enables IVGAE to integrate structural priors and improve imputation robustness under different missingness settings. In addition, a Transformer-based heterogeneous embedding module efficiently encodes categorical features without relying on high-dimensional one-hot representations.

Extensive experiments on 16 real-world datasets show that \texttt{IVGAE} achieves consistently superior performance over existing baselines across all missingness mechanisms, with statistically significant gains in both reconstruction accuracy and downstream classification.

Our contributions are summarized as follows:
\begin{itemize}
    \item We design a \textbf{graph-based structured imputation} framework that represents tabular datasets as bipartite graphs, capturing sample–feature interactions and higher-order dependencies.
    \item We develop a \textbf{dual-decoder variational graph autoencoder} that jointly reconstructs features and infers missingness, enabling mechanism-aware learning of structural priors.
    \item We introduce a \textbf{Transformer-based heterogeneous embedding module} for efficient encoding of categorical features without one-hot expansion.
    \item We conduct a \textbf{comprehensive evaluation} on 16 real-world datasets, showing consistent improvements in imputation accuracy and downstream performance.
\end{itemize}

\section{Related Work}

Data imputation has been extensively studied across multiple paradigms, ranging from statistical heuristics to deep generative and representation learning approaches. This section reviews these three families of methods, highlighting their assumptions, strengths, and limitations, and positions our work in relation to them.

\subsection{Statistical Approaches for Imputation}
Early imputation strategies relied primarily on statistical or model-based heuristics. Simple techniques such as mean, mode, or $k$-nearest neighbors (KNN) imputation~\cite{mean2,kim2021statistical} remain popular due to their ease of use, but they operate independently per feature and ignore inter-feature dependencies. More sophisticated variants, including Multiple Imputation by Chained Equations (MICE)~\cite{mice} and MissForest~\cite{missforest}, attempt to capture non-linear relationships through iterative regression or ensemble learning. Low-rank expectation–maximization (EM) models~\cite{sportisse2020estimation,sportisse2020imputation} further incorporate structural priors by assuming that the complete data matrix lies in a low-dimensional subspace. Nevertheless, these classical techniques are typically limited by strong distributional assumptions and deteriorate when data deviate from the Missing Completely At Random (MCAR) mechanism. Domain-specific solutions developed for recommender systems~\cite{marlin2012collaborative,marlin2009collaborative,wang2019doubly} partially address the Missing Not At Random (MNAR) bias, but their design does not generalize beyond specific application contexts.

\subsection{Deep Generative Models for Imputation}
Recent advances in deep generative modeling have enabled more flexible treatment of incomplete data. Adversarial frameworks such as GAIN~\cite{yoon2018gain} and MisGAN~\cite{misGAN} learn to approximate conditional data distributions through a generator–discriminator interplay, while variational methods such as MIWAE~\cite{miwae} and NOT-MIWAE~\cite{notmiwae} employ probabilistic encoders to infer latent representations consistent with observed entries. Although these approaches outperform traditional baselines, they are primarily designed for continuous variables and require extensive preprocessing of categorical features, often via high-dimensional one-hot encodings that induce information loss. Hierarchical models such as HIVAE~\cite{hivae} extend VAEs to heterogeneous data but rely on rigid parametric forms that constrain flexibility. Several recent surveys~\cite{emmanuel2021survey,miaoreivew,sun2023deep,zhang2023systematic} have emphasized that most generative imputers remain agnostic to the structure of missingness itself, which limits their robustness under MAR or MNAR conditions common in real-world tabular datasets.

\subsection{Graph-Based Approaches}
Graph-based methods have emerged as an effective means to capture relational dependencies in data. Early studies such as GC-MC~\cite{gcmc} and IGMC~\cite{igmc} applied graph convolutional architectures to matrix completion, while GraphVAEs were explored for link prediction and graph generation tasks. More recent works, including GRAPE~\cite{grape} and IGRM~\cite{igrm}, represent tabular datasets as bipartite graphs to jointly model sample–feature interactions and global structural relationships. Although these methods achieve competitive results, most treat missing values deterministically and neglect the generative process of missingness itself. As a consequence, their performance degrades in mechanism-driven settings, particularly when MAR or MNAR patterns encode latent structural information.

Our work builds upon this line of research by introducing a mechanism-aware perspective to graph-based imputation. The proposed \textbf{IVGAE} framework employs a \textit{dual-decoder variational graph autoencoder} that reconstructs both features and missingness masks through probabilistic edge modeling. This design enables the model to capture informative relational priors reflective of the underlying missingness mechanism. Furthermore, a Transformer-based heterogeneous embedding module allows IVGAE to jointly model numerical and categorical variables without lossy one-hot transformations. By bridging graph-based representation learning with mechanism-aware inference, IVGAE advances the robustness and flexibility of deep imputation methods in heterogeneous tabular domains.

\section{IVGAE: The Proposed Method}

\subsection{Problem Definition}
We consider a dataset represented by a matrix \( \mathbf{X} \in \mathbb{R}^{n \times p} \), where \( n \) denotes the number of samples and \( p \) the number of features. Let \( \mathcal{X} = \{x_1, \dots, x_n\} \) and \( \mathcal{V} = \{v_1, \dots, v_p\} \) denote the sets of samples and features, respectively, and let \( X_{ij} \) be the value of feature \( v_j \) for sample \( x_i \).

Missingness in the dataset is captured by a binary mask matrix \( \mathbf{M} \in \{0,1\}^{n \times p} \), where \( M_{ij} = 1 \) if \( X_{ij} \) is observed and \( M_{ij} = 0 \) otherwise. The dataset can thus be partitioned into the observed component \( \mathbf{X}^{o} \) and the missing component \( \mathbf{X}^{m} \), determined by \( \mathbf{M} \).

The process that governs how entries become missing is referred to as the \textit{missingness mechanism}. Following Rubin’s taxonomy, three major types are considered:

\begin{itemize}
    \item \textbf{Missing Completely at Random (MCAR)} — The probability of missingness is independent of both observed and unobserved data, i.e., \( f(\mathbf{M}) \). For instance, a sensor may fail sporadically, causing random data loss.
    \item \textbf{Missing at Random (MAR)} — Missingness depends only on the observed data, i.e., \( f(\mathbf{M} \mid \mathbf{X}^{o}) \). For example, older patients may be less likely to complete certain medical surveys.
    \item \textbf{Missing Not at Random (MNAR)} — Missingness depends on the unobserved values themselves, i.e., \( f(\mathbf{M} \mid \mathbf{X}^{m}) \). For example, patients with severe symptoms may deliberately skip follow-up tests.
\end{itemize}

Accurately modeling these mechanisms is essential for reliable imputation. In practice, real-world datasets often exhibit MAR or MNAR patterns, where missingness correlates with observed or latent factors. Consequently, a robust imputation model must not only reconstruct the missing entries in \( \mathbf{X}^{m} \) but also learn the generative process underlying \( \mathbf{M} \).

\subsection{Bipartite Graph Representation}
The dataset is modeled as an undirected bipartite graph \( \mathcal{G} = (\mathcal{X}, \mathcal{V}, \mathcal{E}) \), where \( \mathcal{X} = \{x_1, \dots, x_n\} \) denotes sample nodes and \( \mathcal{V} = \{v_1, \dots, v_p\} \) denotes feature nodes. An edge \( (x_i, v_j) \in \mathcal{E} \) is created when the entry \( X_{ij} \) is observed, i.e., \( M_{ij} = 1 \). Each edge \( e_{ij} \) is assigned a weight corresponding to the observed feature value, such that \( e_{ij} = X_{ij} \). This construction enables the graph to jointly encode sample–feature interactions while preserving the structural dependencies among heterogeneous attributes. The resulting adjacency matrix \( \mathbf{A} \in \mathbb{R}^{n \times p} \) serves as the foundation for learning relational embeddings through graph neural operations.

\subsection{Encoding Heterogeneous Data}
Real-world tabular datasets often comprise both numerical and categorical features, which must be represented in a unified latent space for graph construction. Conventional one-hot encoding expands each categorical feature into multiple binary dimensions, dramatically increasing graph sparsity and computational complexity, particularly for high-cardinality attributes. To overcome this limitation, we employ a \textbf{Transformer-based heterogeneous embedding module} that generates compact, context-aware representations for categorical features. By learning semantic embeddings conditioned on feature co-occurrence patterns, this module effectively reduces dimensionality while maintaining discriminative power and relational consistency across nodes. 

\begin{figure}[t]
    \centering
    \includegraphics[width=0.48\textwidth]{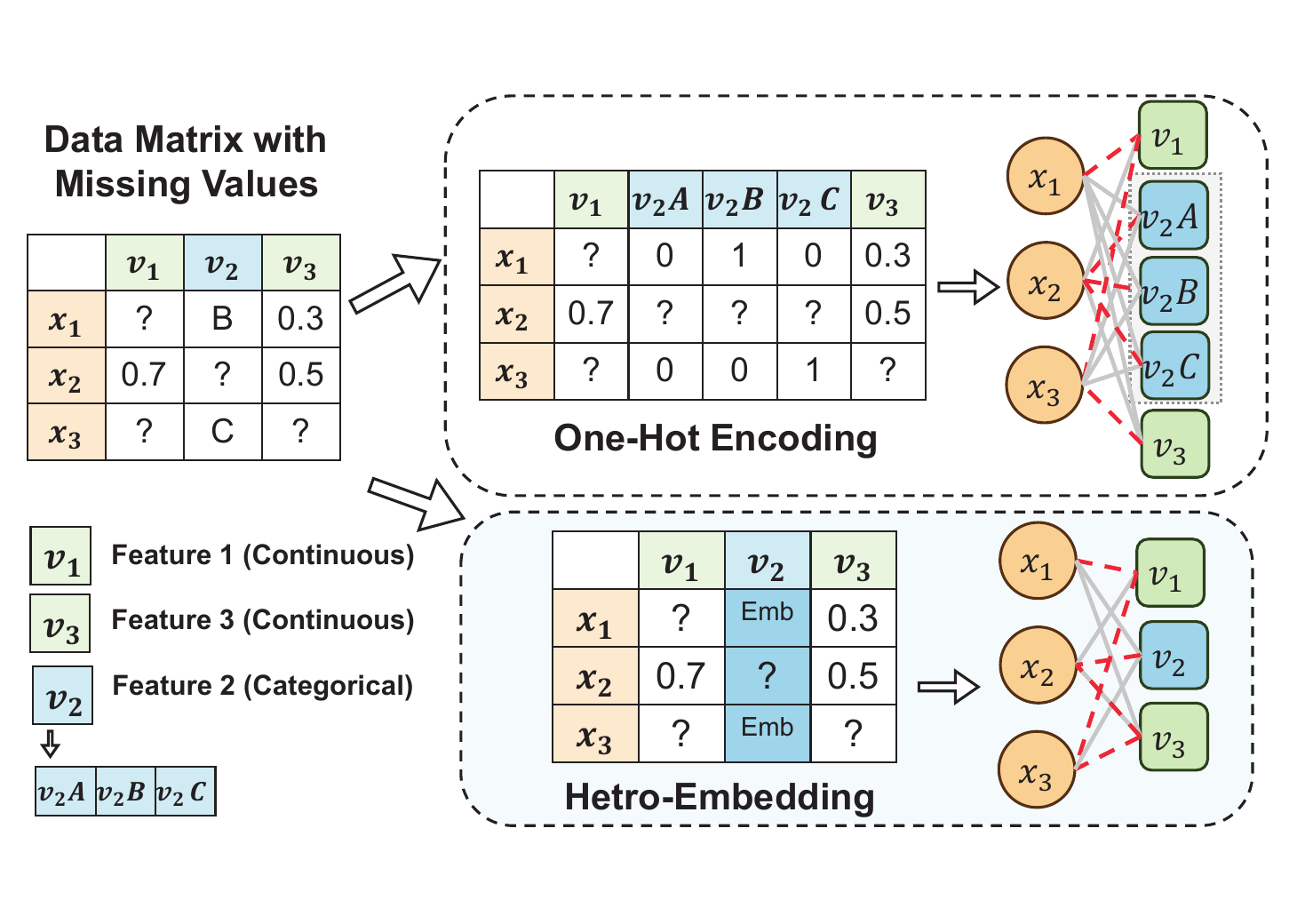}
    \caption{
    Encoding strategies for heterogeneous data within a bipartite graph. 
    \textbf{Top:} One-hot encoding expands each categorical feature (\(F_2\)) into multiple binary nodes, increasing graph size and sparsity. 
    \textbf{Bottom:} The proposed heterogeneous embedding learns compact semantic representations that preserve feature relationships while reducing dimensionality.
    }
    \label{emb}
\end{figure}

As illustrated in Figure~\ref{emb}, one-hot encoding (top) creates a large number of discrete feature nodes, whereas the proposed heterogeneous embedding (bottom) produces dense, low-dimensional vectors that capture semantic similarity among categorical values. This compact representation facilitates efficient graph construction and subsequent message passing within IVGAE.

\begin{figure*}[t]
    \centering
    \includegraphics[width=0.94\textwidth]{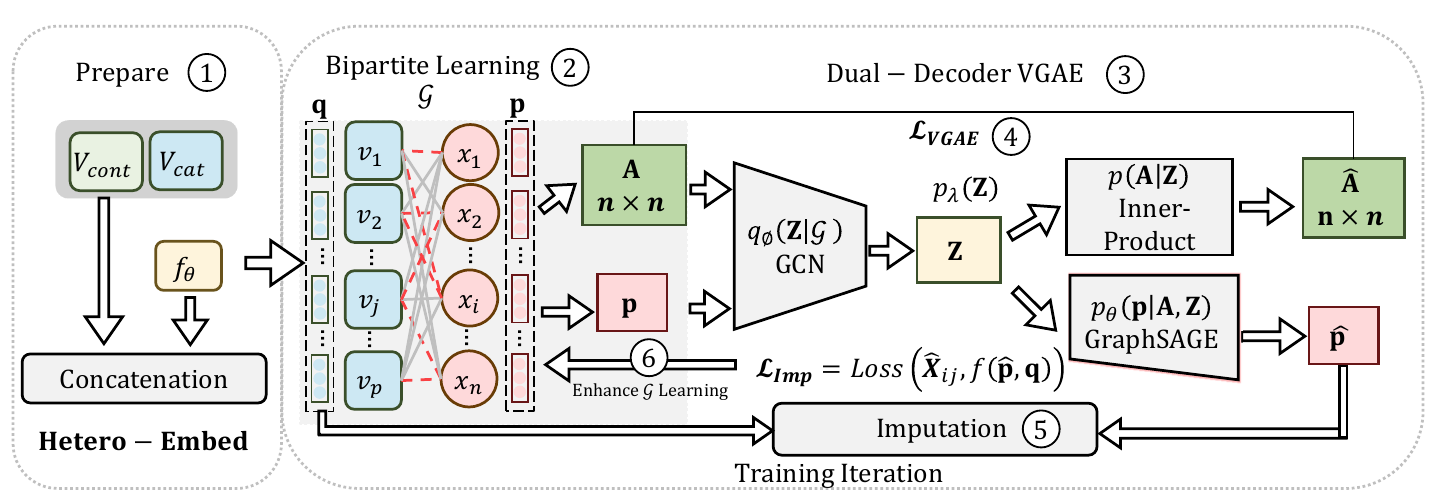}
    \vspace{-0.8em}
    \caption{
    Overview of the proposed IVGAE framework. 
    The model encodes a bipartite graph representation of the dataset, learns latent node embeddings via variational inference, and reconstructs both the feature values \( \hat{\mathbf{X}} \) and the adjacency matrix \( \hat{\mathbf{A}} \) through a dual-decoder mechanism. 
    This design enables simultaneous modeling of feature reconstruction and missingness patterns for mechanism-aware imputation.
    }
    \label{fig:IVGAETraining}
    \vspace{-1.4em}
\end{figure*}

\subsubsection{Heterogeneous Embedding}

To represent heterogeneous tabular features in a unified latent space while accounting for missing values, IVGAE employs a Transformer-based embedding strategy, following prior evidence that attention architectures improve tabular representation learning~\cite{tabtransformer}. Let \( V_{\text{cat}} \) and \( V_{\text{cont}} \) denote the sets of categorical and numerical features, respectively. The embedding process operates on each feature type separately to preserve their statistical characteristics.

For categorical attributes, each feature \( v_{\text{cat}} \in V_{\text{cat}} \) is mapped to a learnable scalar embedding \( e_{\phi_i}(v_{\text{cat}}) \in \mathbb{R} \). A dedicated trainable token is assigned to represent missing categorical values, enabling the model to learn feature-specific patterns even in the presence of incomplete entries. These categorical embeddings are subsequently refined through a stack of Transformer layers \( f_{\theta}(\cdot) \), which model contextual dependencies and interactions among feature tokens:
\begin{equation}
    h_{\text{cat}} = f_{\theta}\big(e_{\phi_i}(v_{\text{cat}})\big),
    \label{eq1}
\end{equation}
where \( h_{\text{cat}} \) denotes the contextualized representation of all categorical features.

\subsubsection{Integration with Numerical Features}

Numerical features \( V_{\text{cont}} \) are standardized to zero mean and unit variance to ensure consistent scaling across attributes. For entries with missing values, we use the mean of observed samples as an initial placeholder to prevent bias propagation during training. The final feature representation for each instance \( x_i \) is obtained by concatenating the transformed categorical embeddings with the normalized numerical features:
\begin{equation}
    h_i = \text{Concat}\big(h_{\text{cat}}, V_{\text{cont}}\big),
    \label{eq2}
\end{equation}
where \( h_i \) serves as the node feature vector for sample \( x_i \) in the bipartite graph.

This heterogeneous embedding mechanism jointly captures categorical dependencies and numerical continuity within a compact representation space. Compared with one-hot encoding, which inflates dimensionality and sparsifies the graph, the proposed Transformer-based encoder learns semantically rich, low-dimensional embeddings that facilitate efficient graph construction and message passing. The embedding and integration processes described in Equations~\ref{eq1} and~\ref{eq2} correspond to the top module of Step~1 in Figure~\ref{fig:IVGAETraining}.

\subsection{Network Architecture}

Graph-based Variational Autoencoders (GraphVAEs) have been widely explored for graph representation learning and missing data imputation~\cite{graphVAE,graphvae2}. However, they are often computationally inefficient for large-scale tabular datasets due to their dense connectivity and high-dimensional latent spaces. In this work, we adopt the Variational Graph Autoencoder (VGAE) framework~\cite{VGAE}, framing imputation as an edge prediction task in a bipartite graph. This formulation leverages adjacency reconstruction to improve imputation quality while maintaining scalability. Figure~\ref{fig:IVGAETraining} illustrates the overall architecture of \texttt{IVGAE}, which consists of two key components: a \textbf{bipartite graph encoder} that captures structured dependencies between samples and features, and a \textbf{dual-decoder VGAE} that reconstructs both feature embeddings and missingness patterns.

Conventional graph-based imputers~\cite{grape} treat missing values as deterministic edge predictions in a bipartite graph \( \mathcal{G} \), estimating them as
\begin{equation}
    \hat{X}_{ij} = \mathcal{F}(\mathcal{G}),
\end{equation}
where \( \mathcal{F} \) denotes a mapping function optimized to minimize reconstruction loss. While effective in learning local relationships, such models fail to explicitly capture the generative process of missingness, leading to suboptimal performance under MAR or MNAR conditions. 
\texttt{IVGAE} addresses this limitation through a \textbf{dual-decoder mechanism} that jointly reconstructs (i) the refined sample embeddings \( \mathbf{p} \) to capture feature dependencies and (ii) the adjacency matrix \( \mathbf{A} \) to explicitly model the missingness structure. This design embeds mechanism-aware priors into the learning process, enhancing robustness to structured missingness.

\subsubsection{Bipartite Graph Learning}

The bipartite graph \( \mathcal{G} = (\mathcal{X}, \mathcal{V}, \mathcal{E}) \) jointly encodes observed feature values and their missingness patterns. Building on GRAPE~\cite{grape} and G2SAT~\cite{g2sat}, IVGAE learns latent node embeddings by aggregating messages from neighboring nodes across multiple graph convolutional layers. At each layer \( l \), the sample node embedding \( \mathbf{p}_i^{(l)} \) is updated based on messages received from connected feature nodes:
\begin{equation}
h_i^{(l)} = \text{Mean}\left(\sigma\left(\mathbf{W}^{(l)} \cdot 
\text{Concat}(\mathbf{q}_j^{(l-1)}, e_{ij}^{(l-1)})\right)\right),
\label{eq4}
\end{equation}
where \( \mathbf{W}^{(l)} \) is a trainable weight matrix, \( e_{ij}^{(l-1)} \) represents the edge embedding at layer \( l-1 \), and \( \sigma(\cdot) \) is a non-linear activation function. 
The operation aggregates feature-level information and propagates it to the corresponding sample nodes, allowing the model to encode both attribute dependencies and graph topology.

\subsubsection{Node and Edge Updates}

At each layer \( l \), sample nodes, feature nodes, and edge embeddings are jointly refined to capture interdependencies between data instances and attributes. Their updates are defined as:
\begin{equation}
\begin{aligned}
    \mathbf{p}_i^{(l)} &= \sigma\left(\widehat{\mathbf{W}}^{(l)} \cdot \text{Concat}(\mathbf{p}_i^{(l-1)}, h_i^{(l)})\right), \\
    \mathbf{q}_j^{(l)} &= \sigma\left(\mathbf{Q}^{(l)} \cdot \text{Concat}(\mathbf{q}_j^{(l-1)}, h_j^{(l)})\right), \\
    e_{ij}^{(l)} &= \sigma\left(\mathbf{U}^{(l)} \cdot \text{Concat}(e_{ij}^{(l-1)}, \mathbf{p}_i^{(l)}, \mathbf{q}_j^{(l)})\right),
\end{aligned}
\label{eq5}
\end{equation}
where \( \widehat{\mathbf{W}}^{(l)} \), \( \mathbf{Q}^{(l)} \), and \( \mathbf{U}^{(l)} \) are learnable transformation matrices for sample nodes, feature nodes, and edges, respectively. These updates enable bidirectional message passing between samples and features, iteratively enriching the learned representations. 

Equations~\ref{eq4}–\ref{eq5} correspond to Step~2 in Figure~\ref{fig:IVGAETraining}, depicting how IVGAE propagates contextual information through the bipartite graph. This hierarchical encoding forms the foundation for the variational inference and reconstruction processes described in the following subsection.
\subsection{Latent Representation Learning via Variational Graph Autoencoder (VGAE)}

To learn expressive representations for imputation, IVGAE adopts the Variational Graph Autoencoder (VGAE)~\cite{VGAE}, which models the distribution of latent node embeddings conditioned on graph structure. Unlike deterministic graph autoencoders, VGAE employs variational inference to capture epistemic uncertainty in the latent space, allowing the model to remain robust under incomplete or noisy observations. Given an initial adjacency matrix \( \mathbf{A} \) that encodes pairwise sample relations derived from the bipartite graph, IVGAE jointly reconstructs two complementary components:

\begin{itemize}
    \item A reconstructed \textbf{adjacency matrix} \( \hat{\mathbf{A}} \), which explicitly models relational dependencies and missingness patterns;
    \item Refined \textbf{sample embeddings} \( \hat{\mathbf{p}} \), which capture high-order feature dependencies informed by the underlying graph topology.
\end{itemize}

The reconstruction of \( \hat{\mathbf{A}} \) and \( \hat{\mathbf{p}} \) allows the model to encode both relational priors and feature-level semantics, thereby improving the consistency and accuracy of downstream imputation.

\subsubsection{Adjacency Matrix and Missingness Mechanisms}

The adjacency matrix \( \mathbf{A} \) captures latent sample similarity, where \( \mathbf{A}_{ij}=1 \) indicates that samples \( x_i \) and \( x_j \) share comparable characteristics~\cite{igrm,VGAE}. Reconstructing \( \hat{\mathbf{A}} \) provides IVGAE with the ability to integrate structural priors that reflect real-world missingness patterns and to propagate reliable information from observed neighbors.

This mechanism is particularly advantageous under non-random missingness conditions.  
For datasets exhibiting \textbf{MNAR}, where missingness depends on unobserved attributes, samples with similar profiles often share comparable missing patterns~\cite{gomer2021subtypes}. By inferring \( \hat{\mathbf{A}} \) through variational learning, IVGAE transfers information across such neighborhoods, mitigating biases caused by systematic data omission.  
In contrast, for \textbf{MAR} scenarios—where missingness correlates with observed features—the reconstructed adjacency structure ensures that inferred values remain consistent with observed relational patterns, thereby improving imputation fidelity.  
Through this mechanism-aware reconstruction, IVGAE effectively bridges sample-level similarity and feature-level dependencies, achieving robust imputation across diverse missingness settings.

\subsubsection{Variational Inference}

Within IVGAE, the encoder learns a probabilistic latent representation of each node conditioned on the observed graph structure. 
Following the formulation of Variational Graph Autoencoders (VGAE)~\cite{VGAE}, the posterior distribution over latent variables is defined as
\begin{equation}
    q_\phi(\mathbf{Z} | \mathbf{p}, \mathbf{A}) = 
    \prod_{i=1}^{N} \mathcal{N}(\mathbf{z}_i | \boldsymbol{\mu}_i, 
    \text{diag}(\boldsymbol{\sigma}_i^2)),
    \label{eq6}
\end{equation}
where each node embedding \( \mathbf{z}_i \) follows a Gaussian distribution parameterized by mean vector \( \boldsymbol{\mu}_i \) and diagonal covariance \( \boldsymbol{\sigma}_i^2 \). 
Both parameters are generated through a shared two-layer Graph Convolutional Network (GCN):
\[
    \boldsymbol{\mu} = \text{GCN}_\mu(\mathbf{p}, \mathbf{A}), \qquad 
    \log \boldsymbol{\sigma} = \text{GCN}_\sigma(\mathbf{p}, \mathbf{A}),
\]
where the GCN operation is defined as
\begin{equation}
    \text{GCN}(\mathbf{p}, \mathbf{A}) 
    = \tilde{\mathbf{A}}\, \text{ReLU}(\tilde{\mathbf{A}}\mathbf{p}\mathbf{W}_0)\mathbf{W}_1,
    \label{eq7}
\end{equation}
with \( \tilde{\mathbf{A}} \) denoting the symmetrically normalized adjacency matrix and 
\( \mathbf{W}_0, \mathbf{W}_1 \) representing learnable transformation matrices.  
Through this formulation, the encoder aggregates structural information from neighboring nodes to infer latent variables that reflect both feature similarity and relational connectivity.  
By introducing stochasticity into node embeddings, the variational framework captures epistemic uncertainty inherent to incomplete data, which improves generalization during imputation.

\subsubsection{Graph Reconstruction}

The decoder in IVGAE comprises two complementary components responsible for reconstructing (i) the adjacency structure and (ii) the node features.

\textbf{Adjacency Reconstruction.}
The first component reconstructs the adjacency matrix \( \hat{\mathbf{A}} \) via an inner-product decoder that predicts the likelihood of an edge between any pair of nodes:
\begin{equation}
    p(\mathbf{A} | \mathbf{Z}) 
    = \prod_{i=1}^{N}\prod_{j=1}^{N} p(\mathbf{A}_{ij} | \mathbf{z}_i, \mathbf{z}_j),
    \qquad
    p(\mathbf{A}_{ij}=1 | \mathbf{z}_i, \mathbf{z}_j)
    = \sigma(\mathbf{z}_i^\top \mathbf{z}_j),
    \label{eq8}
\end{equation}
where \( \sigma(\cdot) \) denotes the sigmoid activation.  
This probabilistic reconstruction enables the model to recover latent relational patterns, effectively performing link prediction that reflects structured missingness within the data.

\textbf{Sample-Embedding Reconstruction.}
The second component refines the node embeddings by reconstructing feature representations through a deterministic GraphSAGE decoder~\cite{GraphSAGE}:
\begin{equation}
    p_\theta(\mathbf{p} | \mathbf{A}, \mathbf{Z}) 
    = \text{GraphSAGE}(\mathbf{Z}, \mathbf{A}).
    \label{eq9}
\end{equation}
For each node \( i \), its reconstructed embedding is obtained by aggregating information from its neighbors:
\begin{equation}
    \hat{\mathbf{p}}_i 
    = \delta \!\left( 
        \mathbf{\Theta} \cdot 
        \text{Concat}\!\left( X_i, 
        \sum_{j \in N(i)} 
        \hat{\mathbf{p}}_j\, \tilde{\mathbf{A}}_{ij} \right)
      \right),
    \label{eq10}
\end{equation}
where \( \mathbf{\Theta} \) is a trainable weight matrix, \( \delta(\cdot) \) denotes a nonlinear activation, and \( N(i) \) is the set of neighbors of node \( i \).  
This decoder aggregates local neighborhood information to produce expressive node representations that preserve both content and structure.  

Equations~\ref{eq6}–\ref{eq10} correspond to Step~3 in Figure~\ref{fig:IVGAETraining}, illustrating how the encoder–decoder pipeline jointly learns probabilistic latent embeddings and reconstructs both graph topology and node features.  
Together, these processes enable IVGAE to capture structural priors and uncertainty for mechanism-aware data imputation.

\subsubsection{Training Objective}

The training of IVGAE is based on maximizing the evidence lower bound (ELBO), which approximates the marginal likelihood of the observed graph structure:
\begin{equation}
    \mathcal{L}_{\text{VGAE}} 
    = \mathbb{E}_{q_\phi(\mathbf{Z} | \mathbf{p}, \mathbf{A})}
      \left[ \log p_\theta(\mathbf{A} | \mathbf{Z}) \right]
      - \text{KL}\!\left[
        q_\phi(\mathbf{Z} | \mathbf{p}, \mathbf{A})
        \,\|\, p(\mathbf{Z})
      \right],
\end{equation}
where the first term encourages accurate reconstruction of the adjacency matrix given latent variables, and the second term regularizes the approximate posterior toward the prior distribution \( p(\mathbf{Z}) \). The Kullback–Leibler (KL) divergence term prevents overfitting by enforcing smoothness in the latent space. The optimization proceeds through full-batch gradient descent using the reparameterization trick~\cite{kingma2013auto}, enabling efficient and differentiable sampling of latent variables.

Although IVGAE employs a dual-decoder architecture, its optimization primarily focuses on reconstructing the adjacency matrix \( \mathbf{A} \), which implicitly guides the imputation of missing values. Since unobserved entries correspond to absent edges in the bipartite graph, accurate adjacency reconstruction encodes structural priors that reflect the underlying missingness mechanism. Variational inference over \( \mathbf{A} \) thus enhances robustness across MCAR, MAR, and MNAR settings by capturing relational uncertainty in the data. Meanwhile, node feature reconstruction \( \mathbf{p} \) is treated deterministically through the GraphSAGE decoder~\cite{GraphSAGE}, which refines local feature representations via neighborhood aggregation without explicit stochastic modeling.

\paragraph*{Comparison with Alternative Graph Generative Models.}
An alternative strategy, GraphVAE~\cite{graphVAE,graphvae2}, directly generates complete graph structures rather than reconstructing edges. 
While this approach can model complex global dependencies, it suffers from two major drawbacks for structured imputation. 
First, the generative process incurs quadratic computational complexity \( O(n^2) \), limiting its scalability for large tabular datasets. 
Second, the resulting graphs require node alignment or graph matching across samples, a process that becomes intractable in heterogeneous feature spaces. 
In contrast, IVGAE circumvents these limitations by framing imputation as an edge prediction problem, which is computationally efficient and inherently aligned with the structure of incomplete tabular data.  
This design preserves scalability while maintaining interpretability in the learned relational priors.
\subsection{Edge Imputation}

The final stage of IVGAE involves imputing missing feature values through edge prediction on the bipartite graph.  
Each missing entry \( X_{ij} \) is estimated using the learned sample and feature embeddings \( \hat{\mathbf{p}}_i \) and \( \mathbf{q}_j \):
\begin{equation}
    \hat{X}_{ij} = 
    \sigma\!\left(
        f\!\left(\text{Concat}(\hat{\mathbf{p}}_i, \mathbf{q}_j)\right)
    \right),
    \label{eq12}
\end{equation}
where \( f(\cdot) \) is a feed-forward neural network that maps the concatenated embeddings into a predicted value, and \( \sigma(\cdot) \) ensures valid numerical ranges.  
This formulation treats imputation as an edge-wise regression problem, enabling the model to infer missing entries by leveraging both sample-to-feature and feature-to-sample relationships learned during graph encoding.

\paragraph*{Loss Formulation for Heterogeneous Data.}
To accommodate different feature types, IVGAE employs a hybrid loss design.  
For datasets using \textbf{heterogeneous scalar embeddings}—where categorical variables are projected into continuous scalar representations—the mean squared error (MSE) is applied:
\[
    \mathcal{L}_{\text{Imp}} = \text{MSE}(\hat{X}_{ij}, X_{ij}),
\]
which encourages smooth reconstruction while preventing over-confident predictions.  
When \textbf{one-hot encoding} is adopted for categorical variables, an auxiliary cross-entropy (CE) term is introduced to ensure discrete category consistency, resulting in the combined objective:
\begin{equation}
    \mathcal{L}_{\text{Imp}} 
    = \text{CE}(\hat{X}_{ij}, X_{ij}) 
    + \text{MSE}(\hat{X}_{ij}, X_{ij}).
    \label{eq14}
\end{equation}
This dual loss stabilizes learning across mixed feature domains, balancing categorical precision with continuous smoothness.

Equations~\ref{eq12}–\ref{eq14} correspond to Step~5 in Figure~\ref{fig:IVGAETraining}, where edge-wise predictions are generated by the dual-decoder module.  
Step~6 depicts iterative refinement of the bipartite graph as reconstructed embeddings progressively improve imputation quality.

\begin{algorithm}[ht]
\caption{IVGAE Training Procedure}
\label{algo1}
\KwIn{Heterogeneous data $\mathbf{V}_{\text{cont}}, \mathbf{V}_{\text{cat}}$, initial adjacency $\mathbf{A}$, number of iterations $T$}
\KwOut{Imputed matrix $\hat{{X}}$, predicted adjacency $\hat{\mathbf{A}}$}

\textbf{Heterogeneous Embedding} \\
$X \leftarrow \text{Concat}(f_\theta(\mathbf{V}_{\text{cat}}),\ \mathbf{V}_{\text{cont}})$
\For{$t \leftarrow 1$ \KwTo $T$}{
    \textbf{Bipartite Graph Encoding} \\
    Construct $\mathcal{G}^{(t)}$ from $X$ and compute node embeddings $\mathbf{p}^{(t)}, \mathbf{q}^{(t)}$ via Eq.~\eqref{eq5} 

    \textbf{Dual-Decoder VGAE Module} \\
    $\mathbf{Z} \leftarrow q_\phi(\mathbf{Z} \mid \mathbf{p}^{(t)}, \mathbf{A})$ using Eq.~\eqref{eq6}, \eqref{eq7}
    
    $\hat{\mathbf{A}}^{(t)} \leftarrow p_\lambda(\mathbf{A} \mid \mathbf{Z})$ using Eq.~\eqref{eq8}
    
    $\hat{\mathbf{p}}^{(t)} \leftarrow p_\theta(\mathbf{p} \mid \mathbf{A}, \mathbf{Z})$ using Eq.~\eqref{eq9}, \eqref{eq10}

    \textbf{Imputation and Loss Computation} \\
    $\hat{X}_{ij} = \mathcal{F}(\mathcal{G})$, using Eq.~\eqref{eq12}
    
    $\mathcal{L}_{\text{imp}} = \text{CE}(\hat{X}_{ij}, X_{ij}) + \text{MSE}(\hat{X}_{ij}, X_{ij})$ using Eq.~\eqref{eq14}

    \textbf{Graph Refinement} \\
    $\mathcal{G}^{(t+1)} \leftarrow \text{Update } \mathcal{G}^{(t)} \text{ via } \hat{\mathbf{p}}^{(t)}$
}
\Return{$\hat{X}, \hat{\mathbf{A}}$}
\end{algorithm}

The algorithm above summarizes the iterative learning procedure of IVGAE.  
At each iteration, updated node embeddings are propagated through the dual-decoder to refine both adjacency reconstruction and imputed values.  
This iterative process ensures convergence toward a stable relational structure, enabling mechanism-aware and type-adaptive imputation across diverse tabular datasets.

\begin{table}[!t]
\caption{Summary of benchmark datasets. ``Cate.'' and ``Cont.'' denote the number of categorical and continuous features, respectively. ``Classes'' indicates the number of target labels for downstream classification tasks, and ``--'' denotes datasets used solely for unsupervised evaluation.}
\label{tab:dataset_summary}
\centering
\resizebox{0.48\textwidth}{!}{%
\begin{tabular}{|l|r|r|r|r|}
\hline
\textbf{Dataset} & \textbf{Samples} & \textbf{Cate.} & \textbf{Cont.} & \textbf{Classes} \\
\hline
Adult        & 48,842  & 7  & 7  & 2  \\ 
Australian   & 690  & 0  & 8  & 2  \\
Banknote     & 1,372   & 0  & 4  & 2  \\
Breast       & 2,869   & 4  & 0  & 2  \\ 
Car          & 1,728   & 6  & 0  & 4  \\ 
Concrete     & 1,030   & 0  & 8  & -- \\ 
Diabetes     & 520     & 15 & 1  & -- \\ 
DOW30        & 2,448   & 0  & 12 & -- \\ 
E-commerce   & 10,999  & 7  & 3  & -- \\ 
Heart        & 1,025   & 9  & 4  & 5  \\ 
Housing      & 506     & 1  & 12 & -- \\ 
Sonar        & 208     & 0  & 60 & 2  \\
Spam         & 4,601   & 0  & 57 & 2  \\ 
Student      & 649     & 11 & 2  & 5  \\ 
Wine         & 4,898   & 0  & 12 & 2  \\ 
Yacht        & 308     & 0  & 6  & -- \\
\hline
\end{tabular}%
}
\end{table}

\section{Experiments}
\subsection{Experimental Setup}
\subsubsection{Datasets}

We evaluate \texttt{IVGAE} using sixteen publicly available datasets collected from the UCI Machine Learning Repository\footnote{\url{https://archive.ics.uci.edu/}} and Kaggle\footnote{\url{https://www.kaggle.com/}}, summarized in Table~\ref{tab:dataset_summary}.  
These datasets cover a broad range of application domains—including finance, healthcare, and sensor data—and exhibit diverse statistical properties such as varying sample sizes, feature dimensionalities, and proportions of categorical versus continuous attributes.  
Such diversity makes them particularly well-suited for benchmarking imputation models designed to handle heterogeneous tabular data.

To simulate incomplete data conditions, we introduce artificial missingness under three mechanisms—\textbf{MCAR}, \textbf{MAR}, and \textbf{MNAR}—at multiple missing rates.  
This controlled setup allows a systematic comparison of imputation robustness under different structural assumptions.  

Model performance is assessed from two complementary perspectives:  
\begin{enumerate}
    \item \textbf{Imputation accuracy}, which quantifies the model’s ability to reconstruct the original data values; and  
    \item \textbf{Downstream predictive performance}, which evaluates how well the imputed data supports subsequent supervised learning tasks (e.g., classification).  
\end{enumerate}
For datasets associated with classification tasks, the number of target classes is indicated in Table~\ref{tab:dataset_summary}, while datasets without explicit labels are used solely to evaluate reconstruction quality.  
This dual evaluation strategy provides a comprehensive assessment of \texttt{IVGAE}’s effectiveness in both generative and task-oriented contexts.

\subsubsection{Missing Data Generation}

In real-world scenarios, the underlying missingness mechanism is typically unknown and difficult to infer, making direct evaluation of imputation methods challenging.  
To ensure reproducibility and controlled benchmarking, we systematically introduce missing values according to three well-defined mechanisms—\textbf{MCAR}, \textbf{MAR}, and \textbf{MNAR}—at missing rates of 10\%, 30\%, 50\%, and 70\%.  
This design enables consistent comparison across methods by applying predefined missingness patterns to complete datasets.

\paragraph*{MCAR (Missing Completely at Random).}
Entries are removed uniformly at random across all features, ensuring that the probability of missingness is independent of both the feature itself and all other variables.  
This setting provides a neutral baseline for assessing imputation under unstructured data loss.

\paragraph*{MAR (Missing at Random).}
The probability of a feature \( x_k \) being missing depends on the values of another observed feature \( x_{k'} \).  
Following the procedure in~\cite{ot}, values in \( x_k \) are removed when \( x_{k'} \) exceeds or falls below a predefined threshold (e.g., the mean or percentile).  
This mechanism mimics practical cases where missingness is conditionally dependent on observed covariates.

\paragraph*{MNAR (Missing Not at Random).}
Missingness depends on the unobserved values of the feature itself~\cite{missmecha}.  
For continuous variables, values within extreme ranges (e.g., top or bottom 10\%) are more likely to be removed, simulating biased omissions of outliers or rare events.  
For categorical variables, specific categories are assigned higher missingness probabilities.  
If the overall missing rate remains below the target level, additional categories are iteratively selected until the desired proportion is reached.  
This mechanism captures self-dependent data omission patterns commonly encountered in domains such as healthcare or finance.

By incorporating all three mechanisms at multiple missing rates, this experimental design provides a comprehensive and reproducible evaluation environment that covers both random and structured data absence scenarios.

\subsubsection{Baseline Methods}

We compare \texttt{IVGAE} against a diverse set of representative imputation baselines spanning statistical, deep generative, and graph-based paradigms.  
This selection ensures a fair and comprehensive comparison across traditional, neural, and structure-aware models.

\paragraph*{Statistical Methods.}
Classical statistical imputers include \texttt{Mean/Mode}, which fills missing entries using the feature-wise mean (numerical) or mode (categorical);  
\texttt{KNN}~\cite{knn}, which estimates missing values from the average of the $K$ nearest neighbors;  
and \texttt{MICE}~\cite{mice}, which performs iterative regression-based updates through chained equations.  
We also include \texttt{MissForest}~\cite{missforest}, a non-parametric random-forest-based iterative imputer that models feature interactions without explicit distributional assumptions.

\paragraph*{Deep Generative Models.}
The deep generative group comprises \texttt{GAIN}~\cite{yoon2018gain}, a GAN-based approach that learns data distributions via adversarial training;  
\texttt{MIWAE}~\cite{miwae}, a variational autoencoder tailored for MAR data;  
\texttt{N-MIWAE}~\cite{notmiwae}, which extends MIWAE to explicitly model MNAR scenarios;  
and \texttt{HI-VAE}~\cite{hivae}, a hierarchical VAE capable of representing heterogeneous variable types.  
These models capture complex nonlinear dependencies but generally require continuous inputs, limiting their direct applicability to categorical features.

\paragraph*{Graph-Based Methods.}
Finally, we include two structure-aware models: \texttt{GRAPE}~\cite{grape}, which encodes tabular data as bipartite graphs for relational imputation, and \texttt{IGRM}~\cite{igrm}, which enhances graph-based reconstruction through friend-network modeling.  
Most existing graph-based imputers, however, treat missing values deterministically and lack explicit modeling of missingness mechanisms.  
For methods unable to process categorical inputs natively, categorical features are converted via one-hot encoding for consistent evaluation.

This diverse set of baselines enables assessment of \texttt{IVGAE} against both traditional and state-of-the-art paradigms, providing insight into its effectiveness across different methodological classes and missingness assumptions.

\subsubsection{Evaluation Metrics}

We assess imputation performance using the \textbf{Average Error (AvgErr)} computed per variable \( v_j \), with metric selection conditioned on the variable type:  
\begin{equation}
\small
\text{AvgErr}(v_j) = 
\begin{cases}
\frac{1}{n} \sum_{i} (X_{ij} - \hat{X}_{ij})^2 & \text{(continuous)} \\
\frac{1}{n} \sum_{i} \|\text{Emb}(X_{ij}) - \hat{X}_{ij}\|^2 & \text{(categorical, emb)} \\
\frac{1}{n} \sum_{i} \text{CE}(\text{OneHot}(X_{ij}), \hat{X}_{ij}) & \text{(categorical, 1-hot)}
\end{cases}
\label{eq:avgerr}
\end{equation}

Here, \( \text{Emb}(X_{ij}) \) denotes the learned scalar embedding of a categorical feature \( X_{ij} \) in heterogeneous embedding settings, and  
\( \text{OneHot}(X_{ij}) \) represents its one-hot encoding when categorical variables are expanded into binary vectors.  
The operator \( \text{CE}(a,b) \) denotes the cross-entropy loss between the true categorical distribution \( a \) and its predicted counterpart \( b \).  
This formulation ensures a consistent error measure across variable types, enabling fair comparison between imputation strategies operating on heterogeneous data.

For downstream evaluation, we further examine the influence of imputation quality on predictive modeling.  
Following prior work, we train an \texttt{XGBoost} classifier on each imputed dataset to evaluate the extent to which imputed features preserve discriminative structure.  
Model performance is estimated using five-fold cross-validation, with an inner five-fold grid search for hyperparameter tuning.  
The final evaluation metric is the \textbf{F1-score}, which balances precision and recall to reflect classification performance on partially recovered data.  
Together, the \texttt{AvgErr} and F1-score provide complementary perspectives—one quantifying reconstruction fidelity and the other measuring utility in downstream prediction tasks.
\begin{figure*}[!t]
    \centering
    \subfloat[MCAR]{%
        \includegraphics[height=0.28\textheight]{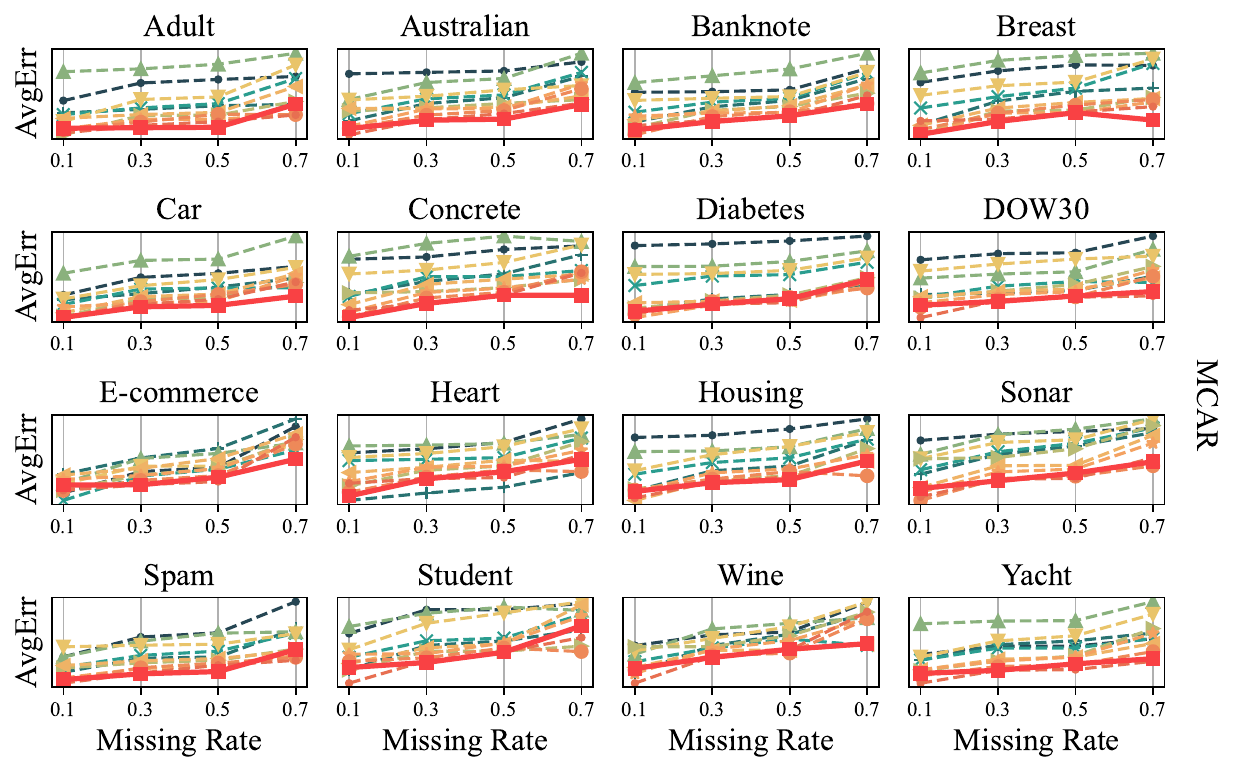}
        \label{fig:mcar}
    }\\[-0.2em]
    \subfloat[MAR]{%
        \includegraphics[height=0.28\textheight]{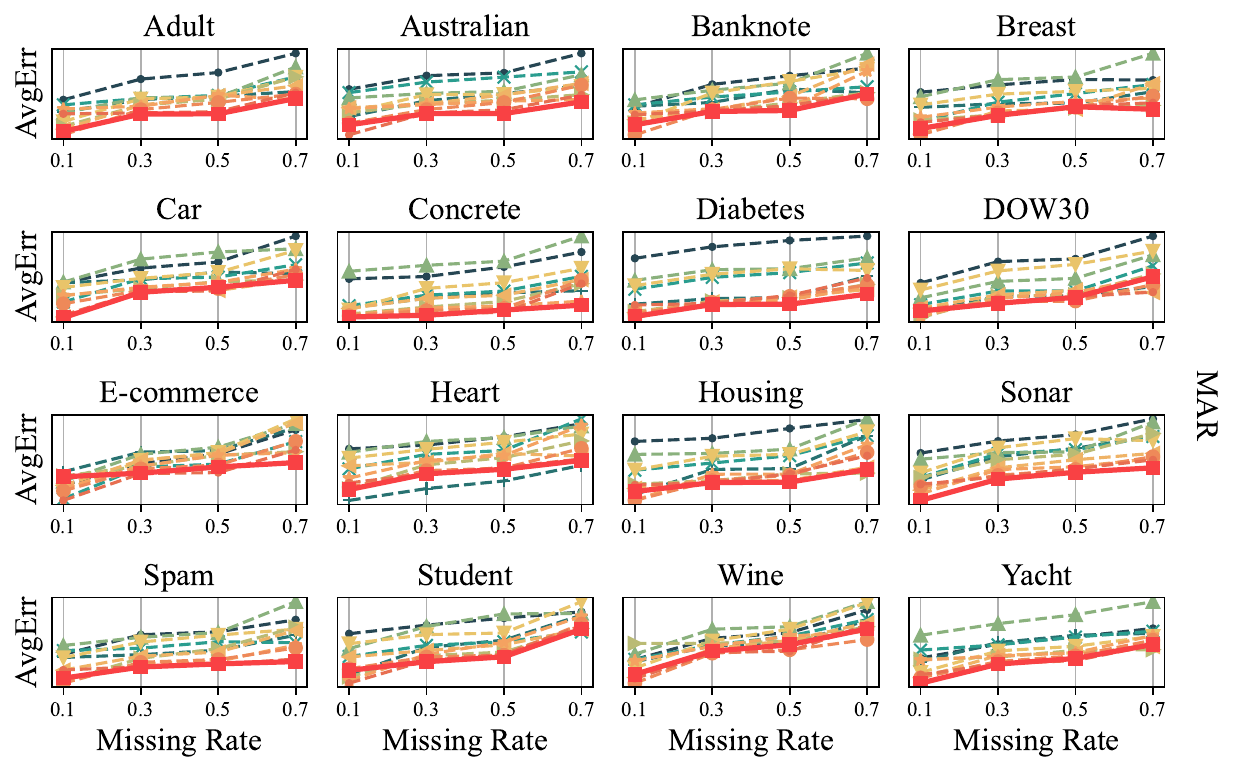}
        \label{fig:mar}
    }\\[-0.2em]
    \subfloat[MNAR]{%
        \includegraphics[height=0.31\textheight]{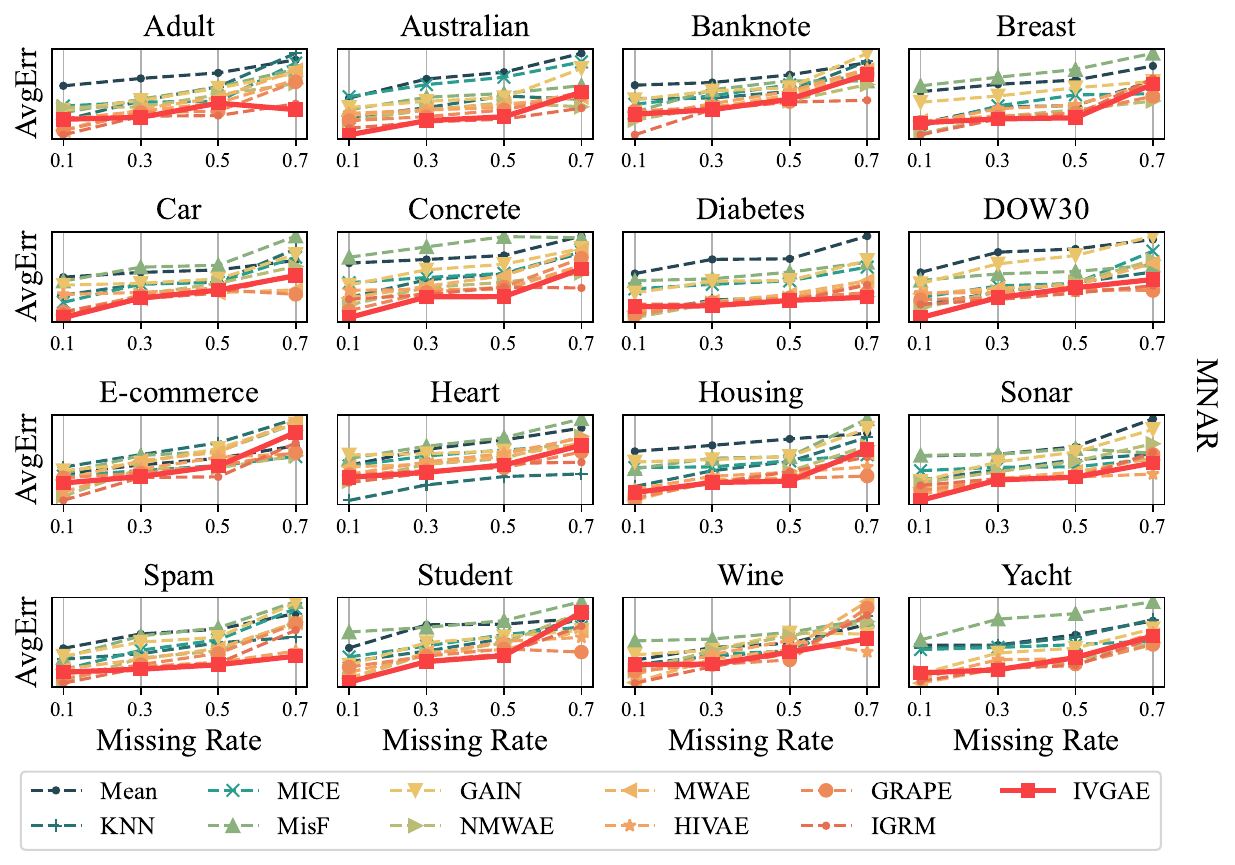}
        \label{fig:mnar}
    }
    \vspace{-0.2em}
    \caption{Comparison of \texttt{AvgErr} across different missingness mechanisms (MCAR, MAR, MNAR). Lower values indicate improved reconstruction accuracy.}
    \label{fig:ratio_results}
\end{figure*}

\begin{table*}[!ht]
\centering
\caption{Comparisons of AvgErr values for missing data imputation across datasets with 30\% MCAR, MAR and MNAR. The best-performing method is \textbf{bolded}, while the second-best is \underline{underlined}. Lower values indicate better performance.}
\label{tab:imputation_comparison_30}
\resizebox{\textwidth}{!}{%
\begin{tabular}{|c|c|c|c|c|c|c|c|c|c|c|c|c|c|c|c|c|}
\hline
\multicolumn{17}{|c|}{\textbf{MCAR}} \\ 
\hline
\textbf{Method} & \textbf{Adult} & \textbf{Aust.} & \textbf{Bank.} & \textbf{Breast} & \textbf{Car} & \textbf{Conc.} & \textbf{Diab.} & \textbf{DOW.} & \textbf{Ecom.} & \textbf{Heart} & \textbf{Hous.} & \textbf{Sonar} & \textbf{Spam} & \textbf{Stud.} & \textbf{Wine} & \textbf{Yacht} \\ 
\hline
\texttt{Mean}   & 0.245 & 0.261 & 0.172 & 0.280 & 0.233 & 0.183 & 0.437 & 0.153 & 0.251 & 0.233 & 0.182 & 0.321 & 0.295 & 0.312 & 0.098 & 0.214  \\ 
\texttt{KNN}    & 0.189 & 0.175 & 0.133 & 0.219 & 0.190 & 0.125 & 0.257 & 0.040 & 0.276 & 0.181 & 0.096 & 0.275 & 0.243 & 0.251 & 0.080 & 0.218  \\ 
\texttt{MICE}   & 0.192 & 0.188 & 0.145 & 0.228 & 0.202 & 0.136 & 0.332 & 0.059 & 0.243 & 0.204 & 0.116 & 0.281 & 0.250 & 0.259 & 0.076 & 0.209  \\ 
\texttt{MisF}   & 0.275 & 0.233 & 0.213 & 0.301 & 0.278 & 0.215 & 0.364 & 0.094 & 0.275 & 0.243 & 0.144 & 0.320 & 0.285 & 0.306 & 0.108 & 0.280  \\ \hline
\texttt{GAIN}   & 0.210 & 0.195 & 0.155 & 0.250 & 0.211 & 0.152 & 0.340 & 0.123 & 0.261 & 0.220 & 0.135 & 0.302 & 0.275 & 0.289 & 0.092 & 0.228  \\ 
\texttt{NMWAE}  & 0.176 & 0.163 & 0.124 & 0.199 & 0.175 & 0.101 & 0.251 & 0.045 & 0.245 & 0.174 & 0.075 & 0.268 & 0.232 & 0.241 & 0.078 & 0.175  \\ 
\texttt{MWAE}   & 0.178 & 0.162 & 0.122 & 0.205 & 0.182 & 0.118 & 0.250 & 0.044 & 0.257 & 0.182 & 0.072 & 0.249 & 0.237 & 0.247 & 0.072 & 0.180  \\ 
\texttt{HIVAE}  & 0.168 & 0.160 & 0.117 & 0.195 & 0.165 & 0.099 & \textbf{0.237} & 0.035 & 0.244 & 0.182 & 0.091 & 0.235 & 0.218 & 0.237 & 0.074 & 0.175  \\ \hline
\texttt{GRAPE}  & 0.162 & 0.143 & 0.108 & 0.195 & 0.169 & 0.086 & 0.249 & 0.020 & \underline{0.235} & 0.155 & 0.076 & 0.220 & 0.220 & 0.230 & 0.063 &  \underline{0.161}  \\ 
\texttt{IGRM}   &  \underline{0.155} & \underline{ 0.131} &  \underline{0.099} & \underline{ 0.183} & \underline{0.161} &  \underline{0.074} & 0.241 &  \underline{0.018} & \textbf{0.227} &  \underline{0.151} &  \underline{0.069} & \underline{0.218} & \underline{0.207} & \underline{0.225} &  \underline{0.062} & \textbf{0.151}  \\  \hline
\texttt{IVGAE}  & \textbf{0.150} & \textbf{0.129} & \textbf{0.095} & \textbf{0.178} & \textbf{0.155} & \textbf{0.073} &  \underline{0.240} & \textbf{0.015} & \textbf{0.227} & \textbf{0.150} & \textbf{0.067} & \textbf{0.215} & \textbf{0.205} & \textbf{0.222} & \textbf{0.060} & \textbf{0.151}  \\ 
\hline\hline
\multicolumn{17}{|c|}{\textbf{MAR}} \\ 
\hline
\textbf{Method} & \textbf{Adult} & \textbf{Aust.} & \textbf{Bank.} & \textbf{Breast} & \textbf{Car} & \textbf{Conc.} & \textbf{Diab.} & \textbf{DOW.} & \textbf{Ecom.} & \textbf{Heart} & \textbf{Hous.} & \textbf{Sonar} & \textbf{Spam} & \textbf{Stud.} & \textbf{Wine} & \textbf{Yacht} \\ 
\hline
\texttt{Mean}   & 0.263 & 0.277 & 0.191 & 0.298 & 0.251 & 0.205 & 0.459 & 0.175 & 0.269 & 0.247 & 0.197 & 0.341 & 0.311 & 0.327 & 0.113 & 0.233  \\ 
\texttt{KNN}   & 0.208 & 0.191 & 0.151 & 0.237 & 0.218 & 0.141 & 0.276 & 0.059 & 0.295 & 0.186 & 0.119 & 0.295 & 0.256 & 0.275 & 0.095 & 0.238  \\ 
\texttt{MICE}   & 0.211 & 0.255 & 0.163 & 0.242 & 0.221 & 0.150 & 0.351 & 0.077 & 0.261 & 0.221 & 0.135 & 0.304 & 0.274 & 0.280 & 0.097 & 0.231  \\
\texttt{MisF}   & 0.213 & 0.216 & 0.178 & 0.315 & 0.276 & 0.238 & 0.378 & 0.109 & 0.290 & 0.257 & 0.159 & 0.307 & 0.304 & 0.324 & 0.135 & 0.297  \\ \hline
\texttt{GAIN}   & 0.210 & 0.212 & 0.172 & 0.268 & 0.222 & 0.170 & 0.362 & 0.144 & 0.276 & 0.235 & 0.152 & 0.321 & 0.291 & 0.305 & 0.107 & 0.212  \\ 
\texttt{NMWAE}  & 0.193 & 0.181 & \textbf{0.131} &  \underline{0.201} & 0.193 & 0.115 & 0.267 & 0.059 & 0.263 & 0.192 & 0.091 & 0.286 & 0.246 & 0.256 & 0.101 & \textbf{0.171}  \\ 
\texttt{MWAE}   & 0.193 & 0.181 & 0.135 & 0.213 &  \underline{0.185} & 0.141 & 0.264 & 0.064 & 0.277 & 0.199 &  \underline{0.090} & 0.262 & 0.252 & 0.268 & 0.088 & 0.178  \\ 
\texttt{HIVAE}  & 0.190 & 0.178 & 0.137 & 0.226 & 0.197 & 0.112 & 0.258 & 0.052 & 0.263 & 0.204 & 0.107 & 0.257 & 0.231 & 0.259 & 0.094 & 0.196  \\ \hline
\texttt{GRAPE}  & 0.181 & 0.161 & 0.135 & 0.207 & 0.188 & 0.105 & 0.263 & 0.041 & \underline{0.257} &  \underline{0.176} & 0.091 & 0.240 & 0.235 & \underline{0.253} & \underline{0.077} & 0.179  \\ 
\texttt{IGRM}   &  \underline{0.172} &  \underline{0.149} &  \underline{0.132} & \underline{0.201} &  \underline{0.185} &  \underline{0.096} &  \underline{0.259} & \underline{0.033} & \textbf{0.247} & \underline{0.176} & \textbf{0.086} &  \underline{0.236} &  \underline{0.223} &  \underline{0.245}& \textbf{0.079} &  \underline{0.172} \\ \hline
\texttt{IVGAE}  & \textbf{0.169} & \textbf{0.147} & \textbf{0.131} & \textbf{0.199} & \textbf{0.183} & \textbf{0.091} & \textbf{0.255} & \textbf{0.036} &  \textbf{0.247} & \textbf{0.167} & \textbf{0.086} & \textbf{0.229} & \textbf{0.222} & \textbf{0.242} & 0.082 & \textbf{0.171}  \\ 
\hline\hline
\multicolumn{17}{|c|}{\textbf{MNAR}} \\ 
\hline
\textbf{Method} & \textbf{Adult} & \textbf{Aust.} & \textbf{Bank.} & \textbf{Breast} & \textbf{Car} & \textbf{Conc.} & \textbf{Diab.} & \textbf{DOW.} & \textbf{Ecom.} & \textbf{Heart} & \textbf{Hous.} & \textbf{Sonar} & \textbf{Spam} & \textbf{Stud.} & \textbf{Wine} & \textbf{Yacht} \\ 
\hline
\texttt{Mean}  & 0.266 & 0.276 & 0.194 & 0.297 & 0.256 & 0.204 & 0.461 & 0.179 & 0.272 & 0.248 & 0.194 & 0.342 & 0.313 & 0.329 & 0.112 & 0.235  \\ 
\texttt{KNN}   & 0.212 & 0.190 & 0.155 & 0.233 & 0.223 & 0.141 & 0.277 & 0.056 & 0.296 & 0.184 & 0.121 & 0.293 & 0.262 & 0.267 & 0.095 & 0.234  \\
\texttt{MICE}  & 0.209 & 0.259 & 0.161 & 0.236 & 0.223 & 0.148 & 0.348 & 0.075 & 0.259 & 0.225 & 0.132 & 0.301 & 0.271 & 0.279 & 0.098 & 0.229  \\ 
\texttt{MisF}  & 0.216 & 0.221 & 0.176 & 0.317 & 0.271 & 0.243 & 0.373 & 0.112 & 0.294 & 0.259 & 0.157 & 0.342 & 0.308 & 0.323 & 0.129 & 0.301  \\\hline
\texttt{GAIN}  & 0.213 & 0.209 & 0.173 & 0.264 & 0.225 & 0.172 & 0.362 & 0.143 & 0.276 & 0.233 & 0.153 & 0.318 & 0.292 & 0.287 & 0.107 & 0.215  \\ 
\texttt{NMWAE} & 0.197 &0.182 & 0.134 & 0.205 & 0.193 & 0.117 & 0.262 & 0.057 & 0.257 & 0.196 & 0.087 & 0.287 & 0.251 & 0.260 & 0.106 & \textbf{0.172}  \\ 
\texttt{MWAE}  & 0.193 & 0.184 & 0.131 & 0.208 & \underline{0.182} & 0.121 & 0.269 & 0.068 & 0.262 & 0.196 & 0.095 &  \underline{0.265} & 0.248 & 0.264 & 0.089 & 0.175  \\ 
\texttt{HIVAE} & 0.187 & 0.183 & 0.141 & 0.228 & 0.199 & 0.111 & \textbf{0.255} & 0.055 & 0.262 & 0.204 & 0.105 & \textbf{0.261} & 0.229 & 0.257 & 0.098 & 0.198  \\ \hline
\texttt{GRAPE} & 0.184 & \underline{0.163} & 0.132 &  \underline{0.205} & 0.189 & 0.102 & 0.266 & 0.038 &  \underline{0.253} & 0.174 & 0.089 & \textbf{0.261} & 0.234 & 0.255 &  \underline{0.078} &  \underline{0.177}  \\ 
\texttt{IGRM } &  \underline{0.175} & \textbf{0.147} & \underline{ 0.127} & \textbf{0.199} & \textbf{0.181} &  \underline{0.095} & \underline{0.259} & \textbf{0.034} & \textbf{0.243} & \textbf{0.166} & \textbf{0.085} &  \underline{0.265} & \underline{0.226} & \underline{0.244} & \textbf{0.074} &  \underline{0.177} \\ \hline
\texttt{IVGAE} & \textbf{0.172} & \textbf{0.147} & \textbf{0.126} & \textbf{0.199} & \underline{0.182} & \textbf{0.088} & \textbf{0.255} & \underline{0.036} & \textbf{0.243} & \underline{0.168} & \underline{0.087} & \textbf{0.261} & \textbf{0.222} & \textbf{0.242} & \textbf{0.074}&  \textbf{0.172} \\ 
\hline
\end{tabular}%
}
\end{table*}

\subsection{Implementation and Parameter Settings}
\texttt{IVGAE} is implemented with a three-layer GraphSAGE encoder (64 hidden units) and a variational latent dimension of 32. 
The adjacency matrix $\mathbf{A}$ is initialized randomly and optimized jointly with model parameters. 
All models are trained for 20{,}000 epochs using the Adam optimizer with a learning rate of 0.001, 
ReLU activations, edge dropout of 0.3, and Min–Max normalization.

\section{Results and Analysis}

\input{figures/cdplot}

\subsection{Qualitative Analysis}

Figures~\ref{fig:ratio_results} and~\ref{fig:cdplot} compare the \texttt{AvgErr} performance of all methods under the three missingness mechanisms (MCAR, MAR, and MNAR) across varying missing rates.  
As expected, imputation accuracy generally declines as the proportion of missing data increases.  
Nevertheless, \texttt{IVGAE} consistently attains lower reconstruction errors and exhibits smoother degradation trends, indicating stronger robustness to increasing incompleteness.  
The CD diagram in Figure~\ref{fig:cdplot} further confirms these results, showing that \texttt{IVGAE} achieves the best overall rank across all datasets, mechanisms, and missing rates, with statistically significant improvements over most competing methods at $\alpha = 0.05$.

This stability can be attributed to two factors: (1) the dual-decoder structure enables explicit modeling of both feature reconstruction and missingness patterns, and (2) the heterogeneous embedding module allows categorical features to be represented compactly without information loss. These design choices mitigate overfitting and improve generalization across datasets with different feature types and sparsity levels.

Due to space limitations, we report detailed numerical results for a representative missing rate of 30\% in Table~\ref{tab:imputation_comparison_30}, which offers a balanced view of practical imputation difficulty and real-world applicability. At this level, \texttt{IVGAE} outperforms all baselines on the majority of datasets, with notable gains on heterogeneous data such as \texttt{Adult}, \texttt{Heart}, and \texttt{Student}. Minor performance fluctuations on \texttt{Wine} and \texttt{Yacht} (MAR) can be attributed to limited sample sizes and weak relational structures, which reduce the expressive capacity of the bipartite graph—an effect also discussed in~\cite{zhou2024missing}.

\subsection{Downstream Task Evaluation}

\subsubsection{Supervised Learning Classification}
To assess the utility of the imputed data for downstream predictive modeling, we evaluate classification performance using the F1-score under a representative 30\% missing rate (Table~\ref{tab:f1_comparison_30}). This setting reflects a realistic level of data incompleteness commonly encountered in practical scenarios. 

Overall, \texttt{IVGAE} achieves superior or competitive performance across all datasets, outperforming conventional and generative baselines in most cases. In particular, the model yields up to \textbf{5\%} improvement in F1-score over graph-based baselines such as \texttt{GRAPE} and \texttt{IGRM}. This improvement highlights the benefit of mechanism-aware edge modeling, which preserves relational information critical to downstream tasks. 

Slight underperformance on the \texttt{Banknote} (MAR) dataset can be explained by its low feature dimensionality and weak inter-sample dependencies, which limit the advantages of graph-based representation learning. Nevertheless, \texttt{IVGAE} maintains high stability and remains competitive across all settings, reinforcing its ability to generalize to diverse data domains and missingness patterns.

\begin{table}[!t]
\caption{Mean Squared Error (MSE) on the Yahoo! R3 dataset. Lower values indicate better performance.}
\label{tab:yahoo_results}
\centering
\resizebox{\columnwidth}{!}{%
\begin{tabular}{|l|l|c|}
\hline
\textbf{Model Category} & \textbf{Model} & \textbf{MSE ($\downarrow$)} \\ 
\hline
\multirow{4}{*}{Statistical} 
    & \texttt{Mean}        & 2.571 $\pm$ 0.001 \\ \cline{2-3}
    & \texttt{KNN}         & 2.124 $\pm$ 0.002 \\ \cline{2-3}
    & \texttt{MICE }       & 2.028 $\pm$ 0.004 \\ \cline{2-3}
    & \texttt{MisForest }  & 1.987 $\pm$ 0.004 \\ 
\hline
\multirow{4}{*}{Generative}
    & \texttt{GAIN    }    & 1.157 $\pm$ 0.007 \\ \cline{2-3}
    & \texttt{MIWAE  }     & 2.055 $\pm$ 0.001 \\ \cline{2-3}
    & \texttt{NMWAE  }     & 0.939 $\pm$ 0.007 \\ \cline{2-3}
    & \texttt{HIVAE  }     & 0.997 $\pm$ 0.004 \\ 
\hline
\multirow{2}{*}{Graph-based}
    & \texttt{GRAPE   }    & 2.001 $\pm$ 0.007 \\ \cline{2-3}
    & \texttt{IGRM    }    & 0.957 $\pm$ 0.005 \\ 
\hline
Proposed
    & \textbf{\texttt{IVGAE}} & \textbf{0.937} $\pm$ \textbf{0.007} \\ 
\hline
\end{tabular}%
}
\vspace{-0.8em}
\end{table}

\subsubsection{Yahoo! R3 Recommendation Task}
We further evaluate imputation performance on the Yahoo! R3 dataset~\cite{marlin2012collaborative}, which contains \textit{MNAR} user ratings for training and \textit{MCAR} ratings for unbiased testing. As reported in Table~\ref{tab:yahoo_results}, deep generative models substantially outperform traditional baselines, highlighting their strength in capturing non-linear dependencies in user–item interactions. 

Among all methods, \texttt{IVGAE} achieves the lowest MSE, indicating its superior capability to reconstruct unobserved preferences. This improvement can be attributed to its dual-decoder mechanism, which jointly models observed relationships and missingness patterns, thereby capturing structural dependencies overlooked by conventional VAEs or GNN-based imputers. These findings suggest that \texttt{IVGAE} is particularly well-suited for recommendation scenarios where missing ratings exhibit non-random, behavior-driven structures.

\begin{table*}[!ht]
    \centering
    \caption{F1-score comparison under 30\% missing rate on classification tasks.The best-performing method is \textbf{bolded}, and the second-best is \underline{underlined}. Higher values indicate better performance. All results are averaged over 5 runs.}
    \label{tab:f1_comparison_30}
    \begin{tabular}{|c|c|c|c|c|c|c|c|c|c|c|c|}
    \hline
    \multicolumn{11}{|c|}{\textbf{MCAR}} \\ 
    \hline
    \textbf{Method} & \textbf{Adult} & \textbf{Aust.} & \textbf{Bank.} & \textbf{Breast} & \textbf{Car} & \textbf{Heart} & \textbf{Sonar} & \textbf{Spam} & \textbf{Stud.} & \textbf{Wine} \\
    \hline
    \texttt{Mean} & 0.245\textsubscript{$\pm$0.005} & 0.613\textsubscript{$\pm$0.004} & 0.762\textsubscript{$\pm$0.004} & 0.428\textsubscript{$\pm$0.001} & 0.438\textsubscript{$\pm$0.002} & 0.177\textsubscript{$\pm$0.005} & 0.691\textsubscript{$\pm$0.001} & 0.698\textsubscript{$\pm$0.005} & 0.199\textsubscript{$\pm$0.005} & 0.796\textsubscript{$\pm$0.004} \\ 
            \texttt{KNN} & 0.241\textsubscript{$\pm$0.004} & 0.652\textsubscript{$\pm$0.005} & 0.772\textsubscript{$\pm$0.004} & 0.412\textsubscript{$\pm$0.005} & 0.404\textsubscript{$\pm$0.002} & 0.181\textsubscript{$\pm$0.004} & 0.721\textsubscript{$\pm$0.005} & 0.725\textsubscript{$\pm$0.002} & 0.198\textsubscript{$\pm$0.001} & 0.872\textsubscript{$\pm$0.003} \\
            \texttt{MICE} & 0.218\textsubscript{$\pm$0.004} & 0.612\textsubscript{$\pm$0.003} & 0.772\textsubscript{$\pm$0.002} & 0.451\textsubscript{$\pm$0.004} & 0.428\textsubscript{$\pm$0.001} & 0.277\textsubscript{$\pm$0.004} & 0.741\textsubscript{$\pm$0.002} & 0.693\textsubscript{$\pm$0.004} & 0.213\textsubscript{$\pm$0.003} & 0.856\textsubscript{$\pm$0.005} \\ 
            \texttt{EM} & 0.227\textsubscript{$\pm$0.002} & 0.480\textsubscript{$\pm$0.002} & 0.771\textsubscript{$\pm$0.005} & 0.451\textsubscript{$\pm$0.002} & 0.374\textsubscript{$\pm$0.003} & 0.281\textsubscript{$\pm$0.001} & 0.663\textsubscript{$\pm$0.002} & 0.515\textsubscript{$\pm$0.003} & 0.174\textsubscript{$\pm$0.002} & 0.690\textsubscript{$\pm$0.005} \\ 
            \texttt{MisF} & 0.242\textsubscript{$\pm$0.004} & 0.651\textsubscript{$\pm$0.004} & 0.711\textsubscript{$\pm$0.001} & 0.416\textsubscript{$\pm$0.002} & 0.374\textsubscript{$\pm$0.004} & 0.228\textsubscript{$\pm$0.005} & 0.698\textsubscript{$\pm$0.005} & 0.569\textsubscript{$\pm$0.001} & 0.166\textsubscript{$\pm$0.005} & 0.823\textsubscript{$\pm$0.002} \\ \hline
            \texttt{GAIN} & 0.242\textsubscript{$\pm$0.004} & 0.655\textsubscript{$\pm$0.003} & 0.761\textsubscript{$\pm$0.001} & 0.413\textsubscript{$\pm$0.003} & 0.369\textsubscript{$\pm$0.001} & 0.281\textsubscript{$\pm$0.004} & 0.605\textsubscript{$\pm$0.004} & \underline{0.835\textsubscript{$\pm$0.005}} & 0.178\textsubscript{$\pm$0.004} & 0.846\textsubscript{$\pm$0.004} \\ 
            \texttt{NMWAE} & 0.245\textsubscript{$\pm$0.002} & \underline{0.658\textsubscript{$\pm$0.005}} & 0.735\textsubscript{$\pm$0.001} & 0.424\textsubscript{$\pm$0.005} & 0.206\textsubscript{$\pm$0.002} & 0.205\textsubscript{$\pm$0.002} & 0.621\textsubscript{$\pm$0.004} & 0.794\textsubscript{$\pm$0.004} & 0.220\textsubscript{$\pm$0.001} & 0.883\textsubscript{$\pm$0.002} \\ 
            \texttt{MWAE} & 0.244\textsubscript{$\pm$0.003} & 0.517\textsubscript{$\pm$0.001} & \textbf{0.810\textsubscript{$\pm$0.001}} & \underline{0.451\textsubscript{$\pm$0.003}} & 0.352\textsubscript{$\pm$0.004} & 0.282\textsubscript{$\pm$0.004} & \textbf{0.762\textsubscript{$\pm$0.001}} & 0.650\textsubscript{$\pm$0.002} & \underline{0.229\textsubscript{$\pm$0.002}} & 0.861\textsubscript{$\pm$0.005} \\ 
            \texttt{HIVAE} & 0.244\textsubscript{$\pm$0.001} & 0.587\textsubscript{$\pm$0.001} & 0.773\textsubscript{$\pm$0.001} & 0.447\textsubscript{$\pm$0.003} & 0.379\textsubscript{$\pm$0.002} & 0.243\textsubscript{$\pm$0.002} & 0.691\textsubscript{$\pm$0.002} & 0.722\textsubscript{$\pm$0.002} & 0.224\textsubscript{$\pm$0.005} & 0.872\textsubscript{$\pm$0.004} \\ \hline
            \texttt{GRAPE} & \underline{0.248\textsubscript{$\pm$0.003}} & 0.574\textsubscript{$\pm$0.005} & 0.777\textsubscript{$\pm$0.004} & 0.441\textsubscript{$\pm$0.004} & 0.412\textsubscript{$\pm$0.002} & \underline{0.289\textsubscript{$\pm$0.002}} & 0.742\textsubscript{$\pm$0.005} & 0.784\textsubscript{$\pm$0.004} & 0.220\textsubscript{$\pm$0.002} & \underline{0.895\textsubscript{$\pm$0.003}} \\ 
            \texttt{IGRM} & 0.239\textsubscript{$\pm$0.005} & 0.649\textsubscript{$\pm$0.003} & 0.770\textsubscript{$\pm$0.005} & 0.451\textsubscript{$\pm$0.005} & \underline{0.445\textsubscript{$\pm$0.005}} & 0.282\textsubscript{$\pm$0.004} & 0.734\textsubscript{$\pm$0.005} & 0.795\textsubscript{$\pm$0.004} & 0.229\textsubscript{$\pm$0.005} & 0.762\textsubscript{$\pm$0.004} \\ \hline
            \texttt{IVGAE} & \textbf{0.250\textsubscript{$\pm$0.001}} & \textbf{0.658\textsubscript{$\pm$0.002}} & \underline{0.781\textsubscript{$\pm$0.002}} & \textbf{0.461\textsubscript{$\pm$0.003}} & \textbf{0.460\textsubscript{$\pm$0.004}} & \textbf{0.289\textsubscript{$\pm$0.003}} & \underline{0.754\textsubscript{$\pm$0.004}} & \textbf{0.857\textsubscript{$\pm$0.003}} & \textbf{0.236\textsubscript{$\pm$0.003}} & \textbf{0.912\textsubscript{$\pm$0.001}} \\ \hline
    \hline
    \multicolumn{11}{|c|}{\textbf{MAR}} \\ \hline
    \textbf{Dataset} & \textbf{Adult} & \textbf{Aust.} & \textbf{Bank.} & \textbf{Breast} & \textbf{Car} & \textbf{Heart}  & \textbf{Sonar} & \textbf{Spam} & \textbf{Stud.} & \textbf{Wine} \\ 
    \hline
    \texttt{Mean} & 0.237\textsubscript{$\pm$0.004} & 0.512\textsubscript{$\pm$0.002} & 0.852\textsubscript{$\pm$0.004} & 0.478\textsubscript{$\pm$0.004} & 0.518\textsubscript{$\pm$0.002} & 0.185\textsubscript{$\pm$0.005} & 0.751\textsubscript{$\pm$0.005} & 0.576\textsubscript{$\pm$0.003} & 0.233\textsubscript{$\pm$0.004} & 0.778\textsubscript{$\pm$0.001} \\ 
            \texttt{KNN} & 0.235\textsubscript{$\pm$0.004} & 0.505\textsubscript{$\pm$0.002} & 0.847\textsubscript{$\pm$0.002} & 0.475\textsubscript{$\pm$0.002} & 0.452\textsubscript{$\pm$0.003} & 0.195\textsubscript{$\pm$0.001} & 0.795\textsubscript{$\pm$0.003} & 0.595\textsubscript{$\pm$0.005} & 0.235\textsubscript{$\pm$0.003} & 0.822\textsubscript{$\pm$0.003} \\ 
            \texttt{MICE} & 0.235\textsubscript{$\pm$0.004} & \underline{0.524\textsubscript{$\pm$0.002}} & 0.846\textsubscript{$\pm$0.002} & 0.471\textsubscript{$\pm$0.003} & 0.510\textsubscript{$\pm$0.003} & 0.216\textsubscript{$\pm$0.003} & 0.774\textsubscript{$\pm$0.002} & 0.635\textsubscript{$\pm$0.004} & 0.237\textsubscript{$\pm$0.003} & 0.852\textsubscript{$\pm$0.003} \\ 
            \texttt{EM} & 0.196\textsubscript{$\pm$0.001} & 0.484\textsubscript{$\pm$0.003} & 0.826\textsubscript{$\pm$0.003} & 0.494\textsubscript{$\pm$0.003} & 0.376\textsubscript{$\pm$0.002} & 0.231\textsubscript{$\pm$0.003} & 0.712\textsubscript{$\pm$0.003} & 0.537\textsubscript{$\pm$0.002} & 0.179\textsubscript{$\pm$0.002} & 0.718\textsubscript{$\pm$0.005} \\ 
            \texttt{MisF} & 0.272\textsubscript{$\pm$0.003} & 0.419\textsubscript{$\pm$0.002} & 0.855\textsubscript{$\pm$0.004} & 0.413\textsubscript{$\pm$0.004} & 0.393\textsubscript{$\pm$0.004} & 0.235\textsubscript{$\pm$0.003} & 0.756\textsubscript{$\pm$0.005} & 0.595\textsubscript{$\pm$0.003} & 0.171\textsubscript{$\pm$0.003} & 0.887\textsubscript{$\pm$0.001} \\ \hline
            \texttt{GAIN} & 0.279\textsubscript{$\pm$0.002} & 0.491\textsubscript{$\pm$0.003} & 0.811\textsubscript{$\pm$0.004} & 0.413\textsubscript{$\pm$0.003} & 0.416\textsubscript{$\pm$0.001} & \underline{0.281\textsubscript{$\pm$0.001}} & 0.689\textsubscript{$\pm$0.004} & 0.866\textsubscript{$\pm$0.003} & 0.184\textsubscript{$\pm$0.002} & 0.832\textsubscript{$\pm$0.003} \\
            \texttt{NMWAE} & 0.294\textsubscript{$\pm$0.003} & 0.478\textsubscript{$\pm$0.001} & 0.815\textsubscript{$\pm$0.002} & 0.485\textsubscript{$\pm$0.003} & 0.356\textsubscript{$\pm$0.002} & 0.213\textsubscript{$\pm$0.004} & 0.721\textsubscript{$\pm$0.004} & 0.515\textsubscript{$\pm$0.003} & 0.219\textsubscript{$\pm$0.002} & \textbf{0.908\textsubscript{$\pm$0.002} } \\ 
            \texttt{MWAE} & 0.223\textsubscript{$\pm$0.005} & 0.443\textsubscript{$\pm$0.002} & 0.884\textsubscript{$\pm$0.004} & \underline{0.523\textsubscript{$\pm$0.005}} & 0.466\textsubscript{$\pm$0.002} & 0.242\textsubscript{$\pm$0.002} & 0.804\textsubscript{$\pm$0.004} & 0.715\textsubscript{$\pm$0.004} & \textbf{0.265\textsubscript{$\pm$0.001}} & 0.885\textsubscript{$\pm$0.002} \\ 
            \texttt{HIVAE} & 0.245\textsubscript{$\pm$0.001} & 0.445\textsubscript{$\pm$0.003} & \underline{0.886\textsubscript{$\pm$0.003}} & 0.518\textsubscript{$\pm$0.003} & 0.483\textsubscript{$\pm$0.003} & 0.245\textsubscript{$\pm$0.004} & 0.764\textsubscript{$\pm$0.002} & 0.635\textsubscript{$\pm$0.005} & 0.229\textsubscript{$\pm$0.001} & 0.771\textsubscript{$\pm$0.002} \\ \hline
            \texttt{GRAPE} & \underline{0.297\textsubscript{$\pm$0.002}} & 0.494\textsubscript{$\pm$0.001} & 0.852\textsubscript{$\pm$0.002} & 0.481\textsubscript{$\pm$0.003} & \underline{0.536\textsubscript{$\pm$0.003}} & 0.247\textsubscript{$\pm$0.002} & \underline{0.842\textsubscript{$\pm$0.001}} & 0.715\textsubscript{$\pm$0.004} & 0.248\textsubscript{$\pm$0.002} & 0.812\textsubscript{$\pm$0.002} \\
            \texttt{IGRM} & 0.280\textsubscript{$\pm$0.004} & \textbf{0.535\textsubscript{$\pm$0.002}} & \textbf{0.887\textsubscript{$\pm$0.004}} & 0.501\textsubscript{$\pm$0.001} & \textbf{0.534\textsubscript{$\pm$0.005}} & 0.276\textsubscript{$\pm$0.003} & 0.743\textsubscript{$\pm$0.002} & \underline{0.871\textsubscript{$\pm$0.003}} & 0.255\textsubscript{$\pm$0.005} & 0.872\textsubscript{$\pm$0.004} \\ \hline
            \texttt{IVGAE} & \textbf{0.303\textsubscript{$\pm$0.002}} & \textbf{0.535\textsubscript{$\pm$0.003}} & 0.852\textsubscript{$\pm$0.002} & \textbf{0.536\textsubscript{$\pm$0.005}} & \textbf{0.534\textsubscript{$\pm$0.001}} & \textbf{0.297\textsubscript{$\pm$0.002}} & \textbf{0.857\textsubscript{$\pm$0.002}} & \textbf{0.827\textsubscript{$\pm$0.002}} & \underline{0.256\textsubscript{$\pm$0.002}} & \underline{0.902\textsubscript{$\pm$0.001}} \\ \hline
    \hline
    \multicolumn{11}{|c|}{\textbf{MNAR}} \\ 
    \hline
    \textbf{Dataset} & \textbf{Adult} & \textbf{Aust.} & \textbf{Bank.} & \textbf{Breast} & \textbf{Car} & \textbf{Heart} & \textbf{Sonar} & \textbf{Spam} & \textbf{Stud.} & \textbf{Wine}  \\ 
    \hline
    \texttt{Mean} & 0.213\textsubscript{$\pm$0.002} & 0.666\textsubscript{$\pm$0.004} & 0.685\textsubscript{$\pm$0.005} & 0.413\textsubscript{$\pm$0.001} & 0.503\textsubscript{$\pm$0.004} & 0.155\textsubscript{$\pm$0.003} & 0.540\textsubscript{$\pm$0.001} & 0.771\textsubscript{$\pm$0.002} & 0.207\textsubscript{$\pm$0.004} & 0.724\textsubscript{$\pm$0.005} \\
            \texttt{KNN} & 0.221\textsubscript{$\pm$0.001} & 0.624\textsubscript{$\pm$0.001} & 0.684\textsubscript{$\pm$0.005} & 0.453\textsubscript{$\pm$0.005} & 0.453\textsubscript{$\pm$0.004} & 0.175\textsubscript{$\pm$0.003} & 0.580\textsubscript{$\pm$0.005} & 0.726\textsubscript{$\pm$0.003} & 0.188\textsubscript{$\pm$0.005} & 0.754\textsubscript{$\pm$0.001} \\ 
            \texttt{MICE} & 0.232\textsubscript{$\pm$0.003} & 0.670\textsubscript{$\pm$0.005} & 0.703\textsubscript{$\pm$0.001} & 0.441\textsubscript{$\pm$0.004} & 0.495\textsubscript{$\pm$0.003} & 0.210\textsubscript{$\pm$0.002} & 0.581\textsubscript{$\pm$0.003} & 0.730\textsubscript{$\pm$0.004} & 0.176\textsubscript{$\pm$0.005} & 0.727\textsubscript{$\pm$0.001} \\ 
            \texttt{EM} & 0.218\textsubscript{$\pm$0.004} & 0.554\textsubscript{$\pm$0.004} & 0.751\textsubscript{$\pm$0.002} & 0.475\textsubscript{$\pm$0.003} & 0.387\textsubscript{$\pm$0.003} & 0.155\textsubscript{$\pm$0.005} & 0.599\textsubscript{$\pm$0.004} & 0.499\textsubscript{$\pm$0.002} & 0.174\textsubscript{$\pm$0.004} & 0.531\textsubscript{$\pm$0.005} \\ 
            \texttt{MisF} & 0.269\textsubscript{$\pm$0.001} & 0.720\textsubscript{$\pm$0.005} & 0.722\textsubscript{$\pm$0.002} & 0.464\textsubscript{$\pm$0.004} & 0.390\textsubscript{$\pm$0.002} & 0.217\textsubscript{$\pm$0.001} & 0.622\textsubscript{$\pm$0.003} & 0.791\textsubscript{$\pm$0.004} & 0.179\textsubscript{$\pm$0.003} & 0.866\textsubscript{$\pm$0.003} \\ \hline
            \texttt{GAIN} & 0.277\textsubscript{$\pm$0.004} & 0.768\textsubscript{$\pm$0.002} & 0.771\textsubscript{$\pm$0.004} & 0.485\textsubscript{$\pm$0.004} & 0.312\textsubscript{$\pm$0.002} & 0.218\textsubscript{$\pm$0.001} & 0.641\textsubscript{$\pm$0.005} & 0.792\textsubscript{$\pm$0.004} & 0.181\textsubscript{$\pm$0.004} & 0.854\textsubscript{$\pm$0.005} \\
            \texttt{NMWAE} & 0.201\textsubscript{$\pm$0.003} & 0.615\textsubscript{$\pm$0.002} & 0.686\textsubscript{$\pm$0.004} & 0.476\textsubscript{$\pm$0.004} & 0.215\textsubscript{$\pm$0.002} & 0.215\textsubscript{$\pm$0.003} & 0.745\textsubscript{$\pm$0.002} & \underline{0.798\textsubscript{$\pm$0.002}} & 0.219\textsubscript{$\pm$0.002} & 0.875\textsubscript{$\pm$0.003} \\ 
            \texttt{MWAE} & 0.235\textsubscript{$\pm$0.005} & 0.636\textsubscript{$\pm$0.003} & \underline{0.846\textsubscript{$\pm$0.003}} & \textbf{0.542\textsubscript{$\pm$0.002}} & 0.484\textsubscript{$\pm$0.004} & 0.214\textsubscript{$\pm$0.003} & 0.767\textsubscript{$\pm$0.003} & 0.759\textsubscript{$\pm$0.002} & 0.205\textsubscript{$\pm$0.002} & 0.873\textsubscript{$\pm$0.005} \\ 
            \texttt{HIVAE} & 0.202\textsubscript{$\pm$0.003} & 0.613\textsubscript{$\pm$0.003} & 0.775\textsubscript{$\pm$0.003} & 0.495\textsubscript{$\pm$0.002} & 0.331\textsubscript{$\pm$0.003} & 0.235\textsubscript{$\pm$0.001} & 0.726\textsubscript{$\pm$0.005} & 0.812\textsubscript{$\pm$0.005} & \underline{0.276\textsubscript{$\pm$0.001}} & 0.891\textsubscript{$\pm$0.004} \\ \hline
            \texttt{GRAPE} & \underline{0.294\textsubscript{$\pm$0.005}} & 0.608\textsubscript{$\pm$0.002} & 0.805\textsubscript{$\pm$0.003} & 0.466\textsubscript{$\pm$0.003} & 0.448\textsubscript{$\pm$0.004} & 0.235\textsubscript{$\pm$0.003} & \underline{0.778\textsubscript{$\pm$0.003}} & 0.799\textsubscript{$\pm$0.001} & 0.265\textsubscript{$\pm$0.003} & \underline{0.920\textsubscript{$\pm$0.002}} \\ 
            \texttt{IGRM} & 0.275\textsubscript{$\pm$0.002} & \underline{0.786\textsubscript{$\pm$0.003}} & 0.815\textsubscript{$\pm$0.004} & 0.459\textsubscript{$\pm$0.004} & \underline{0.503\textsubscript{$\pm$0.002}} & \underline{0.259\textsubscript{$\pm$0.005}} & 0.576\textsubscript{$\pm$0.003} & 0.583\textsubscript{$\pm$0.003} & 0.263\textsubscript{$\pm$0.004} & 0.879\textsubscript{$\pm$0.002} \\ \hline
            \texttt{IVGAE} & \textbf{0.335\textsubscript{$\pm$0.002}} & \textbf{0.802\textsubscript{$\pm$0.003}} & \textbf{0.851\textsubscript{$\pm$0.004}} & \underline{0.511\textsubscript{$\pm$0.003}} & \textbf{0.505\textsubscript{$\pm$0.002}} & \textbf{0.297\textsubscript{$\pm$0.002}} & \textbf{0.782\textsubscript{$\pm$0.002}} & \textbf{0.819\textsubscript{$\pm$0.003}} & \textbf{0.292\textsubscript{$\pm$0.003}} & \textbf{0.936\textsubscript{$\pm$0.003}} \\ \hline
    \end{tabular}
    \end{table*}
    
\begin{table}[!t]
\centering
\caption{Ablation study on key components of \texttt{IVGAE} under a 30\% missing rate. We report F1-score (↑) and AvgErr (↓) under MCAR, MAR, and MNAR mechanisms.}
\label{tab:ablation_full}
\resizebox{\columnwidth}{!}{%
\begin{tabular}{|l|cc|cc|cc|}
\hline
\multirow{2}{*}{\textbf{Method}} & \multicolumn{2}{c|}{\textbf{MCAR}} & \multicolumn{2}{c|}{\textbf{MAR}} & \multicolumn{2}{c|}{\textbf{MNAR}} \\ 
\cline{2-7}
 & \textbf{F1 (↑)} & \textbf{AvgErr (↓)} & \textbf{F1 (↑)} & \textbf{AvgErr (↓)} & \textbf{F1 (↑)} & \textbf{AvgErr (↓)} \\ 
\hline
\texttt{IVGAE}      & \textbf{0.5658} & \textbf{0.146} & \textbf{0.5899} & \textbf{0.166} & \textbf{0.613} & \textbf{0.168} \\ 
GAE                 & 0.5414 & 0.148 & 0.5679 & 0.168 & 0.571 & 0.168 \\ 
MLP                 & 0.5028 & 0.156 & 0.5117 & 0.174 & 0.524 & 0.175 \\ 
\hline
\texttt{Hetero}     & \textbf{0.5658} & \textbf{0.146} & \textbf{0.5899} & \textbf{0.166} & \textbf{0.613} & \textbf{0.168} \\ 
TLF                 & 0.5214 & 0.168 & 0.5701 & 0.185 & 0.601 & 0.181 \\ 
One-hot             & 0.5514 & 0.149 & 0.5624 & 0.169 & 0.597 & 0.180 \\ 
\hline
\end{tabular}%
}
\vspace{-0.8em}
\end{table}

\subsection{Scalability}

We evaluate the runtime scalability of \texttt{IVGAE} under varying sample sizes, feature dimensions, and missing rates using synthetic datasets. All results are plotted on a logarithmic scale to better visualize differences across magnitudes. The evaluation settings are as follows:
\begin{itemize}
    \item \textbf{Sample Size:} $d{=}30$ features, missing rate $30\%$;
    \item \textbf{Feature Dimension:} $n{=}1000$ samples, missing rate $30\%$;
    \item \textbf{Missing Rate:} $d{=}30$ features, $n{=}1000$ samples.
\end{itemize}

\begin{figure}[h]
    \centering
    \includegraphics[width=0.95\linewidth]{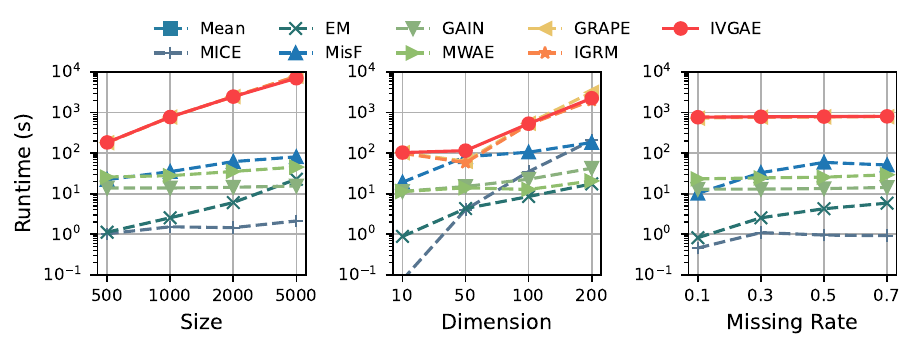}
    \caption{Runtime scalability of \texttt{IVGAE} and representative baselines across varying sample sizes, feature dimensions, and missing rates. 
    The Y-axis is in log scale. Lower values indicate faster execution.}
    \label{fig:runtime_scalability}
\end{figure}

\noindent\textbf{Sample Size.} The runtime of \texttt{IVGAE} increases smoothly with sample size, following a near-linear trend on the log scale. Although deep generative models such as \texttt{GAIN} and \texttt{MWAE} exhibit comparable scaling behavior, their absolute runtime remains higher by one to two orders of magnitude.

\noindent\textbf{Feature Dimension.} As dimensionality grows, graph-based models (\texttt{GRAPE}, \texttt{IGRM}, and \texttt{IVGAE}) show similar scaling characteristics, with no statistically significant difference in runtime. The variance across these methods is minimal, suggesting that message-passing operations dominate computational cost.

\noindent\textbf{Missing Rate.} Across different missing rates, the runtime of \texttt{IVGAE} remains nearly constant, demonstrating robustness to increasing sparsity. Several baseline methods failed to complete at higher missing ratios due to convergence or memory issues, and thus are omitted from the plot. Overall, \texttt{IVGAE} achieves stable and predictable runtime efficiency across all tested conditions.

\subsection{Ablation Study}
\subsubsection{Effect of the Dual-Decoder VGAE}
We evaluate the contribution of variational inference and graph structure learning by comparing the full \texttt{IVGAE} with two reduced variants: (1) a Graph Autoencoder (GAE)~\cite{VGAE} using a deterministic encoder, and (2) a structure-agnostic MLP imputer. Although GAE leverages graph homophily~\cite{zhao2021data}, its deterministic encoding limits the ability to model uncertainty in latent representations. As shown in the upper section of Table~\ref{tab:ablation_full}, replacing the variational module with GAE increases AvgErr by up to 0.03 and reduces F1 by approximately 2–3\% across all missingness mechanisms. The MLP baseline yields the weakest performance, indicating that structural information is crucial for imputing heterogeneous data. These results confirm that both graph-based relational context and variational uncertainty modeling are indispensable for robust imputation.

\subsubsection{Effect of Heterogeneous Data Embedding}
To examine the effectiveness of our heterogeneous embedding (\texttt{Hetero}), we compare it against two alternatives: \texttt{Tree-Driven Latent Factor Encoding (TLF)}~\cite{borisov2023deeptlf} and traditional one-hot encoding. TLF relies on supervised tree-based latent factors, limiting its flexibility in unsupervised or graph-based settings. One-hot encoding, while simple, greatly increases graph sparsity and fails to preserve semantic similarity among categorical values. Specifically, a categorical feature with $C$ categories expands into $C$ binary dimensions, thereby increasing the number of feature nodes and edges in the bipartite graph. For a dataset with $M$ samples and $N_{\text{cat}}$ categorical features, each having an average of $\bar{C}$ categories, the resulting edge count scales as:
\[
|\mathcal{E}_{\text{one-hot}}| = M \cdot (N_{\text{num}} + N_{\text{cat}} \cdot \bar{C}),
\]
which implies a graph complexity of $\mathcal{O}(M \cdot N_{\text{cat}} \cdot \bar{C})$ and leads to a large, sparse adjacency matrix. In contrast, \texttt{Hetero}-Embedding maps each categorical column into a fixed-dimensional latent space, maintaining a compact and semantically meaningful graph representation. This design bounds the number of feature nodes by the original column count, regardless of categorical cardinality. 

As shown in the lower half of Table~\ref{tab:ablation_full}, \texttt{Hetero} consistently achieves lower AvgErr and higher F1, with improvements of up to 10\% over TLF and 4\% over one-hot encoding. Moreover, empirical profiling shows that \texttt{Hetero} reduces inference time by 20–30\% compared to one-hot encoding due to its smaller and denser bipartite structure. For fair comparison, the one-hot variant uses the same loss formulation as Eq.~\ref{eq14}.

\begin{table}[!t]
\centering
\caption{Complexity comparison between encoding strategies. 
$N_{\text{cat}}$ and $\bar{C}$ denote the number and average cardinality of categorical features, 
$M$ the number of samples, and $N$ the total number of features.}
\label{tab:complexity_comparison}
\begin{tabular}{|l|c|c|}
\hline
\textbf{Method} & \textbf{\#Feature Nodes} & \textbf{Edge Count} \\
\hline
One-Hot Encoding & $\mathcal{O}(N_{\text{cat}} \!\cdot\! \bar{C})$ & $\mathcal{O}(M \!\cdot\! N_{\text{cat}} \!\cdot\! \bar{C})$ \\
\texttt{Hetero} & $\mathcal{O}(N)$ & $\mathcal{O}(M \!\cdot\! N)$ \\
\hline
\end{tabular}
\vspace{-0.8em}
\end{table}

\begin{figure}[!t]
    \centering
    \includegraphics[width=\linewidth]{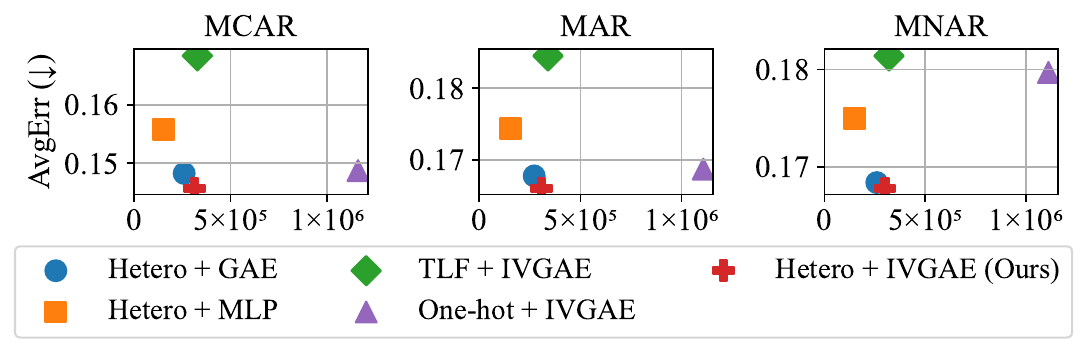}
    \caption{Comparison of imputation error and runtime under a 30\% missing rate, averaged across all datasets. 
    Lower-left points indicate better trade-offs between accuracy and efficiency.}
    \label{fig:runtime}
    \vspace{-0.8em}
\end{figure}

\subsubsection{Runtime Analysis}
Figure~\ref{fig:runtime} illustrates the trade-off between runtime and imputation accuracy under a 30\% missing rate. 
Lower-left points represent preferable performance. 
\texttt{Hetero+IVGAE} achieves the best balance, delivering the lowest imputation error with competitive runtime. 
In contrast, \texttt{One-hot} and \texttt{TLF} incur higher computational costs and larger errors due to their expanded graph representations. 
\texttt{GAE} and \texttt{MLP} variants are faster but substantially less accurate, highlighting the necessity of latent relational modeling. 
These results remain consistent across all missingness mechanisms, confirming the robustness and efficiency of the proposed framework.

\section{Conclusion}
This work introduced \texttt{IVGAE}, a Variational Graph Autoencoder framework for imputing incomplete heterogeneous data. 
By modeling sample–feature dependencies through a bipartite graph and employing a dual-decoder architecture, 
\texttt{IVGAE} jointly reconstructs missing values and missingness patterns. 
A Transformer-based heterogeneous embedding efficiently handles mixed data types without high-dimensional one-hot expansion. 
Comprehensive experiments across 16 real-world datasets demonstrate that \texttt{IVGAE} consistently outperforms existing generative and graph-based methods in both reconstruction accuracy and downstream predictive tasks. 
Nevertheless, the model currently depends on a fixed graph structure, which may limit performance in datasets with weak or noisy feature correlations. 
Future work will explore adaptive graph topology learning and fully generative graph models to further enhance scalability and expressiveness for real-world imputation scenarios.

\section*{Acknowledgment}
This work is supported by the Air Force Office of Scientific Research under award number FA2386-23-1-4003 and Deakin University.

\bibliographystyle{IEEEtran}
\bibliography{refs}


 





\end{document}

%% file: figures/cdplot.tex
\begin{figure}[!t]
    \centering
    \resizebox{0.95\linewidth}{!}{%
\begin{tikzpicture}[
  treatment line/.style={rounded corners=1.5pt, line cap=round, shorten >=1pt},
  treatment label/.style={font=\small},
  group line/.style={ultra thick},
]

\begin{axis}[
  clip=false,
  axis x line=center,
  axis y line=none,
  axis line style={-},
  xmin=1,
  ymax=0,
  scale only axis=true,
  width=\axisdefaultwidth,
  ticklabel style={anchor=south, yshift=1.3*\pgfkeysvalueof{/pgfplots/major tick length}, font=\small},
  every tick/.style={draw=black},
  major tick style={yshift=.5*\pgfkeysvalueof{/pgfplots/major tick length}},
  minor tick style={yshift=.5*\pgfkeysvalueof{/pgfplots/minor tick length}},
  title style={yshift=\baselineskip},
  xmax=11,
  ymin=-6.5,
  height=7\baselineskip,
  xtick={1,3,5,7,9,11},
  minor x tick num=1,
  x dir=reverse,
  title={Critical Difference (All Missing Rate, Average Rank)},
,
]

\draw[treatment line] ([yshift=-2pt] axis cs:{1.21875,0}) |- (axis cs:{0.30208333333333337,-2.5})
  node[treatment label, anchor=west] {IVGAE};
\draw[treatment line] ([yshift=-2pt] axis cs:{1.875,0}) |- (axis cs:{0.30208333333333337,-3.5})
  node[treatment label, anchor=west] {IGRM};
\draw[treatment line] ([yshift=-2pt] axis cs:{3.25,0}) |- (axis cs:{0.30208333333333337,-4.5})
  node[treatment label, anchor=west] {GRAPE};
\draw[treatment line] ([yshift=-2pt] axis cs:{4.78125,0}) |- (axis cs:{0.30208333333333337,-5.5})
  node[treatment label, anchor=west] {HIVAE};
\draw[treatment line] ([yshift=-2pt] axis cs:{5.0625,0}) |- (axis cs:{0.30208333333333337,-6.5})
  node[treatment label, anchor=west] {NMWAE};
\draw[treatment line] ([yshift=-2pt] axis cs:{5.5,0}) |- (axis cs:{11.229166666666666,-7.0})
  node[treatment label, anchor=east] {MWAE};
\draw[treatment line] ([yshift=-2pt] axis cs:{7.0625,0}) |- (axis cs:{11.229166666666666,-6.0})
  node[treatment label, anchor=east] {KNN};
\draw[treatment line] ([yshift=-2pt] axis cs:{7.8125,0}) |- (axis cs:{11.229166666666666,-5.0})
  node[treatment label, anchor=east] {MICE};
\draw[treatment line] ([yshift=-2pt] axis cs:{8.875,0}) |- (axis cs:{11.229166666666666,-4.0})
  node[treatment label, anchor=east] {GAIN};
\draw[treatment line] ([yshift=-2pt] axis cs:{10.25,0}) |- (axis cs:{11.229166666666666,-3.0})
  node[treatment label, anchor=east] {Mean};
\draw[treatment line] ([yshift=-2pt] axis cs:{10.3125,0}) |- (axis cs:{11.229166666666666,-2.0})
  node[treatment label, anchor=east] {MisF};
\draw[group line] (axis cs:{10.25,-1.3333333333333333}) -- (axis cs:{10.3125,-1.3333333333333333});
\draw[group line] (axis cs:{4.78125,-3.6666666666666665}) -- (axis cs:{5.5,-3.6666666666666665});

\end{axis}
\end{tikzpicture}

    }
\caption{Critical Difference (CD) diagram of average ranks based on \texttt{AvgErr} across all datasets, missingness mechanisms (MCAR, MAR, MNAR), and missing rates.  
Lower ranks indicate better imputation performance; methods connected by a horizontal line are not significantly different at $\alpha = 0.05$.}

    \label{fig:cdplot}
\end{figure}